%% file: interaction-arxiv-v2.tex
\documentclass[11pt]{article}

\usepackage[varqu]{zi4}
\usepackage[libertine]{newtxmath} % Keep only newtxmath for math font
\usepackage[top=1in, bottom=1in, left=0.82in, right=0.82in]{geometry}

\usepackage{nicefrac}
\usepackage{hyperref}       % hyperlinks
\usepackage{url}            % simple URL typesetting
\usepackage{booktabs}       % professional-quality tables
\usepackage{amsfonts}       % blackboard math symbols
\usepackage{microtype}      % microtypography
\usepackage{xcolor}         % colors
\usepackage{amsmath}

\usepackage{amsthm}
\usepackage{amsthm}
\usepackage{graphicx}
\usepackage[font=small]{caption}
\usepackage{subcaption}
\usepackage{bm}
\usepackage[ruled,noend]{algorithm2e}
\usepackage{algpseudocode} 
\usepackage{varwidth}
\usepackage{placeins}
\usepackage{mathtools}      % For enhanced math tools
\usepackage{booktabs}       % For tables
\usepackage{verbatim}
\usepackage{enumitem}
\setlist{itemsep=3pt}
\usepackage{soul}
\usepackage{scalerel}
\usepackage{cleveref}
\usepackage{nicefrac}
\usepackage{nicematrix}
\usepackage{csquotes}

\usepackage[numbers,sort]{natbib}

\usepackage{etoolbox}
\newcommand{\ignore}[1]{}
\newtheorem{theorem}{Theorem}[section]
\newtheorem*{theorem*}{Theorem} %unnumbered theorem
\newtheorem{lemma}[theorem]{Lemma}
\newtheorem{corollary}[theorem]{Corollary}
\newtheorem{proposition}[theorem]{Proposition}
\newtheorem{observation}[theorem]{Observation}

\newtheorem{claim}[theorem]{Claim}
\newtheorem{definition}[theorem]{Definition}

\newtheorem*{example*}{Example}

\usepackage{enumitem}
\setlist[enumerate]{leftmargin=*}  % Normalize left margin in enumerate.
\setlist[itemize]{leftmargin=*}        % Normalize left margin in itemize.
\newlist{todolist}{itemize}{2}
\setlist[todolist]{label=$\square$}
\usepackage{pifont}
\newcommand{\seq}[1]{\mathcal{#1}}

\newcommand{\eps}{\varepsilon}
\newcommand{\fscore}[2]{m_{#1}(#2)}
\newcommand{\infseq}[1]{(#1_t)_{t=0}^\infty}
\newcommand{\infsta}[1]{(#1)_{t=0}^\infty}

\newtheorem{tradeoff}{Tradeoff}

\newcount\Comments
\usepackage{xcolor}
\usepackage{cleveref}

\crefname{theorem}{Theorem}{Theorems}
\crefname{lemma}{Lemma}{Lemmas}
\crefname{proposition}{Proposition}{Propositions}
\crefname{tradeoff}{Tradeoff}{Tradeoffs}
\crefname{corollary}{Corollary}{Corollaries}
\crefname{observation}{Observation}{Observations}
\crefname{claim}{Claim}{Claims}

\definecolor{darkgreen}{rgb}{0,0.5,0}
 \usepackage{color-edits}

\addauthor{jk}{blue} %jon
\addauthor{so}{red} %sigal
\addauthor{kd}{purple} %kate
\addauthor{gn}{darkgreen} %gali

\title{AI-Assisted Decision Making with Human Learning\footnote{This is the full version of a paper originally published at EC 2025; DOI: \href{https://doi.org/10.1145/3736252.3742492}{10.1145/3736252.3742492.}}}

\author{%
  Gali Noti \\
  Cornell University \\
  \and
  Kate Donahue \\
  MIT \\
  \and
  Jon Kleinberg \\
  Cornell University \\
	\and
  Sigal Oren \\
  Ben-Gurion University \\
}
\date{}

\begin{document}
% \sloppy
\maketitle

% Abstract. Note that this must come before \maketitle.
\begin{abstract}

\input{abstract2}

\end{abstract}

\maketitle

% Paper body

%%%%%%%%%%%%%%%%%%%%%%%%%%%
\section{Introduction} \label{sec:intro}

\input{intro2}

%%%%%%%%%%%%%%%%%%%%%%%%%%%
\section{Related Literature} \label{sec:literature}

\input{literature}

%%%%%%%%%%%%%%%%%%%%%%%%%%%
\section{Model and Preliminaries} \label{sec:model}

\input{model1}

%%%%%%%%%%%%%%%%%%%%%%%%%%%
\section{Warm-up: Feature Selection with Fixed Human Beliefs} \label{sec:static}

\input{static}

%%%%%%%%%%%%%%%%%%%%%%%%%%%
\section{Feature Selection with Human Learning} \label{sec:learning}

\input{learning1}

%%%%%%%%%%%%%%%%%%%%%%%%%%%
\section{Misspecification} \label{sec:misspecification}

\input{misspecification-full-text}

%%%%%%%%%%%%%%%%%%%%%%%%%%%
\section{Discussion} \label{sec: discussion}

\input{discussion}

%%%%%%%%%%%%%%%%%%%%%%%%%%%

\section*{Acknowledgments} 
This work was supported in part by BSF grant 2018206,  a MIT METEOR fellowship, a Vannevar Bush Faculty Fellowship, a Simons Collaboration grant, and a grant from the MacArthur Foundation.

\clearpage
% \setcitestyle{authoryear}

% Bibliography
\bibliographystyle{plainnat} % ACM-Reference-Format
\bibliography{interaction-updated.bib}

% Appendix
\appendix

\input{appendix}

\end{document}

%% file: abstract2.tex
AI systems increasingly support human decision-making. In many cases, despite the algorithm's superior performance, the final decision remains in human hands. For example, an AI may assist doctors in determining which diagnostic tests to run, but the doctor ultimately makes the diagnosis. This paper studies such AI-assisted decision-making settings, where the human learns through repeated interactions with the algorithm. In our framework, the algorithm -- designed to maximize decision accuracy according to its own model -- determines which features the human can consider. The human then makes a prediction based on their own less accurate model. 

We observe that the discrepancy between the algorithm's model and the human's model creates a fundamental tradeoff: Should the algorithm prioritize recommending more informative features, encouraging the human to learn their importance, even if it results in less accurate predictions in the short term until learning occurs? Or is it preferable to forgo educating the human and instead select features that align more closely with their existing understanding, minimizing the immediate cost of learning? Our analysis reveals how this trade-off is shaped by both the algorithm’s patience (the time-discount rate of its objective over multiple periods) and the human’s willingness and ability to learn. 
We show that optimal feature selection has a surprisingly clean combinatorial characterization, reducible to a stationary sequence of feature subsets that is tractable to compute.  As the algorithm becomes more ``patient'' or the human's learning improves, the algorithm increasingly selects more informative features, enhancing both prediction accuracy and the human's understanding.

%% file: intro2.tex
AI systems are increasingly used to assist humans in making decisions. In many situations, although
the algorithm has superior performance, it can only assist the human by providing recommendations, leaving the final decision to the human, as regulations and ethical principles require that humans remain accountable for consequential choices. For example, 
algorithms assist medical doctors in assessing patients' risk factors and in targeting health inspections and treatments (\cite{musen2021,garcia2019,tomavsev2019,jayatilake2021}), and assist judges in making pretrial release decisions, as well as in sentencing and parole determinations (\cite{courts2020, nyc_cja_2020,compas2019}). However, the final decision ultimately remains at the discretion of the doctors or judges (\cite{casey2019rethinking,european2021proposal}). 
This paper studies such AI-assisted decision-making scenarios where the human decision-maker learns from experience in repeated interactions with the algorithm. In our setting, an algorithm assists a human by 
telling them which information they should use for making a prediction, with the goal of improving the {\em human's} prediction accuracy.

Consider, for instance, a doctor assessing a patient's risk of a bacterial infection. To improve their diagnosis, the doctor can order tests that reveal the values of unknown variables. However, the number of tests the doctor can run is limited (e.g., due to costs or time constraints), and they rely on an algorithm to determine which tests to perform. The algorithm, trained on vast amounts of data, is more accurate than the doctor in estimating the statistical relationships between the unknown variables and the disease risk. The doctor has their own (potentially inaccurate) beliefs about how test results relate to risk assessment and will interpret the results according to their beliefs. The doctor orders the tests recommended by the algorithm
-- either because they recognize its superior data processing capabilities or because an insurance provider conditions funding on following the algorithm's selections. The algorithm's objective is to select the tests that lead the doctor to make the most accurate prediction.

While we phrase the problem in terms of a doctor and an algorithm, similar problems appear in other domains as well. For example, in bail decisions, a judge may use an algorithm to determine which aspects of a defendant’s record to scrutinize. In hiring, an algorithm may guide which aspects of an applicant's file to review or suggest questions for the interview. Similarly, an algorithm may assist investors in due diligence by highlighting key aspects of a firm to review before making an investment. 
Beyond these domains, in which features correspond to tests of different types, a similar challenge arises in product design, e.g., 
when designing dashboards and deciding on a fixed subset of features to display to assist humans in various prediction tasks. 
The choice prioritizes features that the human can interpret correctly over those that may be more useful but are harder to interpret. For a concrete example, many weather apps display humidity which people in general know how to interpret instead of displaying the dew point, even though the latter better captures the discomfort caused by humidity (\cite{NOAA_DewPoint}).

In the remainder of this introduction, we begin with a qualitative discussion that identifies two fundamental tradeoffs in AI-assisted decision-making. We then provide a brief summary of our model and demonstrate how both tradeoffs manifest within our framework through a  concrete example. Finally, we summarize the main results of our formal analysis.

\vspace{5pt} \noindent  \textbf{Fundamental Tradeoffs in AI-Assisted Decision Making.}
A key question an algorithm faces when assisting a human decision-maker is determining what information will be {\em useful to the human}. If the algorithm were making the prediction on its own, the answer would be clear: it would use all available information to maximize accuracy. When access to information is constrained (e.g., due to costs associated with acquiring additional observations), the algorithm would prioritize the most informative data, selecting the observations expected to contribute most to prediction accuracy. However, when the algorithm does not make the prediction on its own but instead  selects information to assist a $human$ in making a prediction, it must also account for the human's ability to use that information correctly.  
The algorithm seeks to balance increasing {\em informativeness} while providing information where the human's and the algorithm's interpretations of the data {\em diverge} less. Our starting point is identifying this tradeoff, which we call the ``informativeness vs. divergence'' tradeoff.

\begin{tradeoff} \label{tradeoff1}
    {\em The Informativeness vs. Divergence Tradeoff:} When selecting information for human use, the algorithm faces a tradeoff between selecting the most informative data and selecting data that minimizes divergence %with lower divergence  
    between the human's model and the algorithm's model of the ground truth. 
\end{tradeoff}

This tradeoff implies that the algorithm may not always select the most informative features for the prediction at hand, but may instead choose less-informative features that fit the human's level of understanding. For example, a medical algorithm might recommend that a doctor performs a less-accurate throat culture, which the doctor has used frequently and knows how to interpret, rather than a newer, more precise blood test for specific proteins that the doctor does not yet know how to interpret its results.
The need to balance informativeness and divergence is fundamental to any human-algorithm interaction where the algorithm provides information to optimize human performance. Recent research reflects a growing awareness of this challenge.
In chess, for instance, skill-compatible AI is designed so that a powerful algorithm playing alongside a less-skilled human does not only choose the best move but one that the human can understand and build upon (\cite{hamade2024}). 
\cite{xu2024persuasion} considers the question of when an algorithm should delegate a decision to a human and what information it should provide. They find that more information does not always lead to better decisions.

 A second fundamental question in human-algorithm interactions arises when moving beyond a single interaction. In this repeated-interaction setting, the human naturally learns from experience by repeatedly making predictions. 
When considering human learning, the algorithm faces a tradeoff between optimizing for short-term versus long-term outcomes: should it optimize for the human's performance in the present, or should it guide them to learn more toward better performance in the future? 
Here, the dilemma of whether to provide more informative or less divergent data is amplified. 
If the algorithm chooses to optimize based on the human's current understanding (e.g., selecting the less-accurate throat culture which the human already knows how to use), doing so repeatedly comes at the risk of preventing learning opportunities and sustaining incorrect beliefs, which may lead to worse performance in the long run. Conversely, if the algorithm chooses to provide information that encourages learning (e.g., instructing the doctor to use the new blood test), it can improve long-term outcomes, but at the cost of an initial learning phase  during which the human may make errors while adjusting their beliefs. We call this the ``fixed vs. growth mindset'' tradeoff.

\begin{tradeoff} \label{tradeoff2}
     {\em The Fixed vs. Growth Mindset Tradeoff:} 
     When a human decision-maker learns through repeated interactions with an algorithm, the algorithm faces a tradeoff between 
     maximizing immediate performance 
     based on the human's current beliefs and 
     fostering long-term performance 
     by leading  
     the human to learn and improve their beliefs over time.
\end{tradeoff}

The fixed vs. growth mindset tradeoff naturally arises in 
various situations where one teaches another person. 
The teacher can rely on the student's current knowledge to achieve the best performance in the short term. Alternatively, the teacher can instill a new and better skill, which may take longer to master but ultimately leads to improved performance and a stronger skill set in the long run.
This tradeoff is related to the classic question of ``giving a fish vs. giving a fishing rod.''\footnote{This tradeoff is also {reminiscent} of the ``sneakers vs. coaching'' metaphor (\cite{hofman2023a}) for thinking about types of AI-Assistance. While ``coaching'' aligns closely with the growth mindset, ``sneakers'' differs from the fixed mindset: in the fixed {mindset} approach, the algorithm does not provide any additional assistance to the human, but rather optimizes solely with their existing knowledge and abilities.}  Providing a fish ensures immediate success but does not contribute to long-term skill development. In contrast, teaching someone to fish requires an initial investment of time and effort but ultimately equips them with the ability to succeed even more in the future -- whether by catching more fish or developing greater self-sufficiency.

Note that neither of these two tradeoffs has a clear ``correct'' answer. The fixed vs. growth mindset tradeoff (Tradeoff \ref{tradeoff2}), as we will see, depends on the human's ability to learn and the time preferences the algorithm was set to optimize for. In cases of emergency, where immediate outcomes are critical, such as helping a patient in urgent need, it may be best to optimize based on the human's current abilities, even if they are capable of learning. By contrast, situations where time constraints are less strict present an opportunity to focus on skill development and long-term growth. 

The informativeness-divergence tradeoff (Tradeoff \ref{tradeoff1}), which may lead the algorithm to provide suboptimal information
according to 
the human's level of understanding, raises the question of whether the algorithm is merely simplifying the problem into a form the human can comprehend or actively manipulating them. While there is no formal distinction between these two cases -- since in both, the algorithm adapts to human limitations at the cost of being less informative -- we tend to perceive them quite differently depending on the context. For example, it seems reasonable to introduce a doctor to a new blood test (low stake scenario), but undesirable when the algorithm's selections lead a doctor to order costly or invasive tests that are less informative than other available options (high stake scenario).

\vspace{5pt} \noindent  \textbf{Model Summary.}
Consider a human who needs to predict the outcome of a variable $y$. The true outcome is given by a function of $n$ features: $y = f(x)$, where the features $x = (x_1,...,x_n)$ are independent random variables that are standardized to have zero mean and unit variance.
We assume the {widely used} linear functional form:\footnote{ Note that this is similar to the linear regression that 
organizations (e.g., hospitals or city governments) often use to weigh factors that human experts (e.g., doctors or judges) are considering when making decisions (e.g., \cite{nyc_cja_2020, blatchford2000risk, alur2024integrating, jones2024linear}).
For instance, \cite{nyc_cja_2020} describes New York City's criminal justice system use of linear models to combine defendant features. It emphasizes transparency (for judges and affected individuals) as a key principle in model building, instead of using more complex ML models. }
 $y = f(x) = c + \sum_{i=1}^n a_i x_i$.

\begin{itemize}
 \item The human's belief about the coefficient of feature $i$ at time $t\geq 0$ is $h_{i,t}$, where {$h_0 = (h_{1,0},h_{2,0}, ..., h_{n,0})$} is their initial belief. Their belief about the constant term is $\bar c$.
 \item The algorithm's estimate of the true coefficients is $a' = (a'_1,...,a'_n)$ and its estimate of $c$ is $c'$. The algorithm's estimates of the human's initial belief is denoted by {$h'_0 = (h'_{1,0},h'_{2,0},...,h'_{n,0})$} and $\bar c'$.
 \item At every time step $t\geq 0$,
 \begin{itemize}
\item The algorithm selects a subset of features $A_t\subseteq[n]$, with $|A_t| \leq k$, where $k\leq n$ is a budget parameter.
\item The human and the algorithm observe the realization of the features in $A_t$.
\item The human makes a prediction $\bar c + \sum_{i \in A_t} h_{i,t} x_i$ and exhibits a loss according to the Mean Squared Error (MSE) of their prediction. We denote this loss by $MSE(A_t,h_t)$.
\item The human updates $h_{i,t+1}$ for all observed features $i\in A_t$ according to some arbitrary learning rule that converges to the true $a_i$ as the number of times that feature $i$ is selected goes to infinity. 
 \end{itemize}
\end{itemize}

The algorithm selects a sequence  of feature subsets,  $\seq{S}=(A_t)_{t=0}^\infty$,  {aiming} 
to minimize the discounted loss of the human's prediction over time: $\sum_{t=0}^\infty \delta^t MSE(A_t,h_t)$, where $\delta \in (0,1)$ is a discounting parameter that was chosen by the entity deploying the algorithm. We also refer to $\delta$ as the ``patience'' parameter: the higher $\delta$ is, the more patient the algorithm is in considering future outcomes. For the majority of the paper, we analyze the case where the algorithm's model of the ground truth is correct (i.e., $a'=a$ and $c'=c$) and the algorithm knows the human's initial beliefs (i.e., $h'_0 = h_0$, and $\bar c' = \bar c $) and the human's convergence rate. Given these assumptions, the algorithm has all the information required for optimizing this discounted loss. 
We provide more details about our model and elaborate on our modeling choices in Section \ref{sec:model}.

As we will see, many of our results are driven by two quantities: the magnitude of a coefficient $a_i$ representing the {\em informativeness} of feature $i$ and the distance between $a_i$ and $h_{i,0}$ representing the {\em divergence} for that feature at time $0$.

\vspace{5pt} \noindent  \textbf{Example.}
To build intuition about our model and how the fundamental tradeoffs arise within this framework, let us revisit our medical diagnosis example in a single-decision setting. Suppose that the doctor has three possible tests they can run, with true coefficients $a_1=0.3$, $a_2=0.2$, and $a_3=0.1$. That is, test $1$ is the most informative, followed by test $2$, and test $3$ is the least informative. However, the doctor overestimates the importance of tests $1$ and $3$, with $h_1=0.8$ and $h_3=0.15$, while accurately interpreting test $2$, with $h_2=a_2=0.2$. Suppose that the algorithm knows both the ground truth and the human's model and can select any subset of tests (i.e., there is no budget constraint, $k=n=3$).

Table \ref{tbl:example-mse} summarizes the MSE of the human's prediction for each of the $2^3$ possible subsets of tests (i.e., features) that the algorithm can select. 
As can be seen, the best human performance (i.e., the least MSE) is achieved by selecting features $2$ and $3$, and therefore this is the algorithm's optimal feature selection in the single-decision setup. 
This optimal subset includes test $2$, which the human perfectly understands ($h_2 = a_2$).
Test $3$ is also selected, despite some divergence between the human's belief and the ground truth ($h_3 \neq a_3$), because the divergence is small enough relative to its informativeness to still make it beneficial. 
Notably, the optimal subset does not include the most informative test (test 1), as the high divergence in the human's interpretation of this test ($h_1 \gg a_1$) outweighs its informativeness. 
  {In Section \ref{sec:static}, we analyze the exact condition under which 
  the algorithm selects features for an optimal subset in a single prediction instance.} 
This illustrates Tradeoff \ref{tradeoff1}: when minimizing the MSE loss of the human's prediction, the algorithm balances between high informativeness and low divergence of the selected features.

\begin{table}[t]     
    \centering
    \renewcommand{\arraystretch}{1.5}
    \begin{tabular}{|c|c|c|c|c|c|c|c|c|}
        \hline
        Feature Subset          & \(\emptyset\) & \(\{1\}\) & \(\{2\}\) & \(\{3\}\) & \(\{1,2\}\) & \(\{1,3\}\) & \(\{2,3\}\) & \(\{1,2,3\}\) \\ \hline
        % MSE             & 0.14         & 0.3       & 0.1       & 0.17      & 0.26        & 0.33        & 0.13        & 0.29         \\
        MSE             & 0.14         & 0.3       & 0.1       & 0.1325      & 0.26        & 0.2925        & \textbf{0.0925}        & 0.2525         \\ \hline
    \end{tabular}
        \caption{\normalfont Mean Squared Error (MSE) for all possible subsets of features  for the example in Section \ref{sec:intro}. The example has three features $\{1,2,3\}$, the algorithm's model of the true coefficients is $a'=a=(0.3, 0.2, 0.1)$, the algorithm's model of the human's coefficients is $h'=h=(0.8, 0.2, 0.15)$, and there is no limit on the number of features that can be selected (i.e., $k=n=3$). }
        \label{tbl:example-mse}
\end{table}

This example also demonstrates the importance of modeling the human's decision-making process in addition to modeling the ground truth. A na\"{\i}ve algorithm, which does not model human decisions but instead bases its feature selection solely on its own estimates, 
would recommend considering all features to minimize the error from its own perspective.  
In our example, this would result in almost the worst possible error, as shown in Table \ref{tbl:example-mse}.

Now, consider the scenario in which the doctor repeatedly interacts with the algorithm to make predictions for a sequence of different patients, and learns from the repeated interaction. Suppose that the {doctor} is a very fast learner, such that after using a test once, they learn its true coefficient for all subsequent predictions. If the algorithm selects all the tests in the first prediction, it incurs a loss of $0.2525$ at that time step. However, with repeated selections of all tests in subsequent steps, the error drops to zero.
By contrast, if the algorithm repeatedly selects only tests $2$ and $3$ (which were optimal in the single-decision scenario), it initially incurs a smaller loss of $0.0925$, but in subsequent predictions, it incurs a loss of $0.09$. That is, the error improves due to learning test $3$, but only to a suboptimal result of $0.09$ in each prediction. The reason is that the human doctor never had the opportunity to learn test $1$. Thus, if the algorithm weighs each repetition equally, for three steps or more, it is better off enduring the initial learning period and allowing the human to make mistakes and improve their {model} 
over time. 
The choice between these two sequences depends on how the algorithm weighs short-term losses versus long-term losses. 
When learning is more gradual, the learning phase lasts longer and has higher costs, which, along with the weight assigned to future outcomes, influence the algorithm's choice as well.
This captures Tradeoff \ref{tradeoff2}: the algorithm trades off the value of teaching the human (the ``growth'' mindset) vs. helping the human perform as best as they can with their current beliefs (the ``fixed'' mindset). The contrast between the algorithm's choices in the single-decision and the learning scenarios demonstrates the importance of taking human learning into account when considering human decision-making in repeated interactions.

\vspace{5pt} \noindent  \textbf{Summary of Analytical Results.}
We analyze the interaction between algorithmic assistance and a learning human decision-maker. {Recall that the algorithm selects a sequence of feature subsets with the objective of minimizing the discounted loss of the human's prediction.}
We begin by characterizing optimal sequences of feature selections. Initially, one might suspect that feature selection is a hard problem due to the large search space: {exponential in the single-interaction setting and unbounded in the repeated-interaction setting.}
Our analysis reveals a surprisingly clean combinatorial structure for this problem. In \cref{thm:stationary}, we show that there exists an optimal sequence that is a stationary sequence -- a sequence in which the algorithm consistently selects the same subset of features at each step (see Section \ref{sec:stationary}). This insight allows us to restrict our focus to stationary sequences, reducing the problem to a finite space, though still exponential in the number of features $n$. Then, we show in \cref{thm:complexity-learning} that for a given value of $\delta$, it is possible to compute an optimal stationary sequence in $\Theta(n \log n)$ time (see Section \ref{sec:complexity}). Moreover, we find that across the full range of $\delta \in (0,1)$, the total number of stationary sequences that can be optimal is at most $\Theta(n^2)$ (\cref{prop-delta-bounded}, Section \ref{sec:tradeoff-analysis}). Notably, our analysis imposes only a mild restriction on human learning, specifically that it satisfies a natural convergence property (see Section \ref{sec:model-human}). 
For the full details of our analysis, see Section \ref{sec:learning}.

Following these results, we focus our attention on optimal stationary sequences and study the conditions under which the algorithm selects more informative feature subsets, and how this choice is influenced by the time preferences in the algorithm's objective function and the efficiency of the human's learning. 
First, holding human learning at a fixed rate, we show that as the algorithm's patience parameter $\delta$ increases, it increasingly selects more informative feature subsets (\cref{prop-delta-efficient}). This improves both prediction accuracy and the human's understanding of the world in the long term. Additionally, we show that there always exists a sufficiently large $\delta$ value above which the algorithm's optimal selection is the most informative feature set (\cref{prop-delta-efficient}). Second, we fix $\delta$ and vary the efficiency of the learning rule. We show that as the human learns more efficiently, the informativeness of the feature set selected by the algorithm increases (\cref{prop-phi-increases}). Moreover, our analysis highlights that it is more beneficial for the human to invest in learning during earlier time steps rather than later ones, as this allows the algorithm to select more informative features and enables the human to extract greater benefits from the interaction with the algorithm.  For the full analysis, see Section \ref{sec:tradeoff-analysis}.

Finally, in Section \ref{sec:misspecification}, we study the impact of errors in the algorithm's knowledge of the ground-truth coefficients and its models of the human's coefficients and learning rate. Roughly speaking, we translate these modeling errors into the maximum possible error in a quantity that we later denote as the \emph{value} of a feature and is used to select an optimal feature set. We show that this maximum possible error quantifies a level of tolerance to algorithmic modeling errors: when the gaps between feature values are large there is a wide error tolerance range and when they are small the impact of suboptimal choices that the algorithm makes is small. 
This behavior results from the structure of algorithmic assistance, in which the human makes the actual predictions, and the algorithm's role is limited to selecting the feature sets for the human to use.

%% file: literature.tex
Our work is situated in the literature on designing algorithms for assisting human decision-makers (e.g., \cite{bansal2021does,greenwood2024designing,benCSCW,tschandl2020,gomez2025human, peng2024no}). 
In particular, we consider a setting in which the algorithm selects for the human 
which features to use and learn for making the best prediction as part of a repeated interaction. 

The majority of the literature assume that the algorithm has direct access to %all the 
information and can give the human a decision recommendation \cite{benFAT,albright2019} or display the human the relevant information for making the decisions (e.g., \cite{dikmen2022effects,du2022role}). A notable exception {in regard to the algorithm's informational structure} is \cite{iakovlev2024value} that theoretically studies a reverse setting, complementary to ours, where humans have discretion to choose, based on situational information, which features to use at each time step, and algorithmic tools are obliged to use the same features.

Empirical papers in this area of AI-assisted human decision-making study the ability of human decision makers to correctly rely on the algorithm \cite{benFAT, benCSCW, albright2019, tschandl2020, buccinca2021trust, zhang2020}. Typically, they do not consider human learning.
Two exceptions are \cite{noti2023learning} that showed experimentally the advantage of an algorithm to not always provide a recommendation and showed evidence that human decision makers learn through repeated interaction with the algorithm, and  \cite{buccinca2024} that present an experimental study that applies reinforcement learning in repeated decisions with algorithmic advice. 
A complementary experimental study, \cite{he2025human}, finds that people systematically overestimate the alignment between their own preferences and generative AI behavior across diverse economic decision-making tasks,  suggesting that humans may tend to follow AI-generated choices and may learn from them.

Our work contributes to the growing literature on designing algorithms that interact with changing human agents. 
In a position paper, \citet{dean2024accounting} call for further development of \enquote{formal interaction models} between algorithms and humans who change over time. 
Topics in this literature include work on human-algorithm collaboration in multi-armed bandits, where the goal is to jointly identify some best arm
(e.g. \cite{chan2019assistive, bordt2022bandit}). Additionally, some works on recommendation systems take into account the fact that human preferences may change over time (e.g. \cite{Agarwal2022DiversifiedRF, agarwal2023online}). 
\citet{tian2023towards} studies a dynamical human-robot interaction setting, where the human's mental model of the robot changes over time. 
{Performative prediction \cite{perdomo2020performative} and learning in Stackelberg games (e.g., \cite{haghtalab2024calibrated}) set up a Stackelberg game to study how to make predictions when people respond to them in a way that shifts the data distribution used for the prediction.}

Also related are works on human-AI teams showing that to increase the overall performance of the team, the AI should take the human model into account and make sure its choices are understandable for the human (chess, \cite{hamade2024}, the video game overcooked \cite{carroll2019utility}). Moreover, it is not always the case that a more accurate algorithm \cite{bansal2021most, bansal2019} or one that provides more information \cite{xu2024persuasion} 
are better.
Our findings highlight that algorithms need to balance between using features that the human understands and features that the human needs to learn to reach better long-term outcomes.
This tension is related to the broader field of explainable or interpretable machine learning (XAI), (e.g., \cite{samek2017explainable, finkelstein2022explainable, heuillet2021explainability} and the survey \cite{arrieta2020explainable}) that aims to explain to a human what the model has learned and develop techniques for explaining the model's predictions \cite{singh2020explainable, bach2015pixel},
or to learn problem representations that are more easily interpretable by the human 
\cite{hilgard2021learning,mahendran2015understanding,nguyen2016multifaceted, nahumdecongestion}.   

Finally, recent works on learning in strategic environments show that %competition or 
misaligned incentives can lead to counterintuitive preferences over informativeness or model complexity \cite{jagadeesan2023improved, handina2024understanding, feng2022bias}, which may seem related to the informativeness-divergence tradeoff. However, while these works consider misaligned strategic agents who learn, in our setting the incentives of the algorithm and the human are aligned, only the human is learning, and 
the technical questions that we consider are very different from these works. E.g., \cite{jagadeesan2023improved, handina2024understanding} are about competition, whereas we consider an algorithm and a learner who are not competing, and \cite{feng2022bias} focuses on the size of the underlying model class, while our optimization is over a fixed class of models.

%% file: model1.tex
In this section, we provide additional details about the model introduced in Section \ref{sec:intro}. 
Recall that we consider a human tasked with predicting the outcome of a variable $y$. The true outcome is given by a linear function of a set of $n$ features $x = \{x_1, ..., x_n\}$,  such that $y = c + \sum_{i=1}^n a_i x_i$, where the coefficients $a_i$ are non-zero. %and distinct. %(i.e., for all $i,j$, $a_i \neq a_j$).
The features $x_i$ are independent random variables drawn from distributions $F_1,\ldots,F_n$ with known means and finite standard deviations. Without loss of generality, throughout our analysis we assume the features are standardized (such that they have zero mean and unit variance; see Appendix \ref{app:general}). The human and the algorithm interact repeatedly as described in Section \ref{sec:intro}.
%For convenience, 
We divide the next discussion into two perspectives: the human's perspective and the algorithm's perspective.

\subsection{The Human} \label{sec:model-human}

At each time step $t$, the human observes the realization of the features $A_t$ that the algorithm selected and makes a prediction $\bar c + \sum_{i \in A_t} h_{i,t} x_i$ (predicting the mean, which is zero, for any unobserved features). This minimizes the Mean Squared Error (MSE) from the human's perspective. As the MSE is a function of $A_t$ and the human's coefficient vector $h_t$, we denote it by $MSE(A_t,h_t)$ and get:

\begin{claim} \label{clm:huma-mse}
    $MSE(A_t,h_t) = (c-\bar c)^2 + \sum_{i \notin A_t} {a_i}^2 + \sum_{i \in A_t} (a_i - h_{i,t})^2 $
\end{claim}

\begin{proof}
We have
    \begin{align*}
MSE(A_t,h_t) &=\mathbb{E}[\big(c + \sum_{i=1}^n a_i x_i - \bar{c} - \sum_{i \in A_t} h_{i,t} x_i\big)^2] \\
%&= (c-\bar c)^2 + \mathbb{E}[\big(\sum_{i=1}^n a_i x_i - \sum_{i \in A_t} h_{i,t} x_i\big)^2] \\
&= (c-\bar c)^2 + \mathbb{E}[\big(\sum_{i=1}^n a_i x_i - \sum_{i \in A_t} h_{i,t} x_i\big)^2] \\
&= (c-\bar c)^2 + \mathbb{E}[(\sum_{i=1}^n a_i x_i)^2]
+ \mathbb{E}[(\sum_{i \in A_t} h_{i,t} x_i)^2]
- 2\,\mathbb{E}[(\sum_{i=1}^n a_i x_i)(\sum_{i \in A_t} h_{i,t} x_i)]
\end{align*}
Observe that
\begin{align*}
 \mathbb{E}[(\sum_{i \in A_t} h_{i,t} x_i)^2] = \sum_{i \in A_t} \sum_{j \in A_t} \mathbb{E}[h_{i,t} h_{j,t} x_i x_j] 
\end{align*}
Since features are independent we have that for $j\neq i$, $\mathbb{E}[x_ix_j]=0$ and since the variance is normalized to $1$, we have that $\mathbb{E}[x_i^2] =1$. Hence, $\mathbb{E}[(\sum_{i \in A_t} h_{i,t} x_i)^2] = \sum_{i \in A_t} h_{i,t}^2$. Similarly, $\mathbb{E}[(\sum_{i=1}^n a_i x_i)^2] = \sum_{i=1}^n a_i^2$ and $2\,\mathbb{E}[(\sum_{i=1}^n a_i x_i)(\sum_{i \in A_t} h_{i,t} x_i)] = 2\sum_{i \in A_t} a_ih_{i,t}$. Putting this together, we get that:
\begin{align} \label{eq:MSE}
    MSE(A_t,h_t) = (c-\bar c)^2 + \sum_{i \notin A_t} {a_i}^2 + \sum_{i \in A_t} (a_i - h_{i,t})^2 
\end{align}
as required.
\end{proof}

Recall that after the human observes a realization of a feature, they learn %about the true coefficient 
and update their coefficient of the feature according to some learning rule. Basic properties that are natural in a learning setting include: 
\begin{itemize}
       \item Initial beliefs: At the beginning of the interaction, the human starts with some initial beliefs.
    \item Improvement with experience: 
    With additional observations, the human's beliefs become closer to the true values.
    \item Asymptotic learning: with infinite observations, the human's beliefs converge to the truth.
\end{itemize}

Formally, in our context, $h_{i,0}$ denotes the human's initial belief about feature $i$ and let $\fscore{i}{t}$ denote the number of times $i$ was selected until time $t$.
The human's beliefs at time $t$, $h_{i,t}$, is a function of $\fscore{i}{t}$. 
The sequence $|h_{i,t} - a_i|$ is decreasing in $\fscore{i}{t}$, and $\lim_{\fscore{i}{t} \rightarrow \infty}  |h_{i,t} - a_i| = 0$. 
Note that since the variables are independent it is reasonable to assume that the learning process is also independent for each variable and hence we make this assumption.

The following definition of $\phi$-convergence of learning functions captures the above properties. 
\begin{definition} \label{def:convergence}
	Let $\phi: \mathbb{N} \to [0,1]$ be a monotone decreasing function with $\phi(0)=1$, $\lim_{m \rightarrow \infty} \phi(m) = 0$. 
    A learning dynamic is {\em $\phi$-convergent} if for every $t$:
$	%\begin{align*}
		(a_i-h_{i,t})^2 = \phi(\fscore{i}{t})\cdot(a_i-h_{i,0})^2.
$	%\end{align*}  
\end{definition}

In our analysis, we consider human learning in the general sense of $\phi$-convergence. This abstracts away the exact mechanism that leads to learning, i.e., what feedback the human receives and how they use this feedback to update their model.

\subsection{The Algorithm}
The objective of the algorithm is determined by its designer who cares about minimizing the loss of the human's prediction. 
The designer sets a budget \( 0 \leq k \leq n \) on the number of features that the human can use for prediction. The limitation to $k<n$ features may arise, for example, from a cost associated with revealing features' values.

Given that we are dealing with an infinite stream of losses, it is natural to account for the timing of each loss and apply an appropriate discount. Here, 
we adopt the widely used exponential discounting approach, where future losses are consistently discounted according to a parameter \( \delta \in (0,1) \). 
Putting this together we have that the objective of the algorithm is to select a sequence of feature subsets $\seq{S}=(A_t)_{t=0}^\infty$ such that $|A_t|\leq k$ for every $t\geq 0$, that minimizes the discounted loss of the human's prediction:
\begin{align} \label{alg:objective-function}
 L(\seq{S}=(A_t)_{t=0}^\infty,\phi,h_0) &= \sum_{t=0}^\infty \delta^t  MSE(A_t,h_t) =  
 \sum_{t=0}^\infty \delta^t \Big( (c-\bar c)^2 + \big(\sum_{i \notin A_t} {a_i}^2 + \sum_{i \in A_t} (a_i - h_{i,t})^2 \big) \Big) 
 \notag \\
  &= \sum_{t=0}^\infty \delta^t \Big((c-\bar c)^2 + \big(\sum_{i \notin A_t} {a_i}^2 + \sum_{i \in A_t} \phi(\fscore{i}{t})(a_i - h_{i,0})^2 \big)\Big)
\end{align}

As \( \delta \) approaches \( 1 \), the designer places greater emphasis on future losses, whereas smaller values of \( \delta \) indicate a stronger preference for minimizing immediate losses. Thus, \( \delta \) can be interpreted as a ``patience'' parameter. 
Note that such an infinite-horizon discounted loss can also represent situations with an 
uncertain interaction length, where the interaction ends in each round with probability \( 1 - \delta \). In this case, an interaction occurring \( t \) steps in the future has a probability of \( \delta^t \) of taking place and is therefore discounted by that factor.

We are mainly interested in settings where the algorithm is, on average, more accurate than the human. Hence, we primarily analyze the case where the algorithm has accurate coefficients (i.e., \( a' = a \) and $c'=c$), and accurate estimates of the human's initial coefficients (\( h'_0 = h_0 \) and $\bar c' = \bar c$), and the \( \phi \) governing their learning dynamics ($\phi'=\phi$, where $\phi'$ is the algorithm's estimate of $\phi$). This means that the algorithm has all the required information to choose a sequence minimizing \( \sum_{t=0}^\infty \delta^t  MSE(A_t,h_t) \). 
To simplify notations, we omit the prime notation from the algorithm's estimates.
In Section \ref{sec:misspecification}, we consider algorithms that have inaccurate model of the human or of the ground truth.
As we will see, the structure of the problem -- where the algorithm does not make the prediction directly but instead selects the features on which the human will base their prediction -- limits the impact of errors that the algorithm may have in modeling both the human and the ground truth.

%%%%%%%%%%%%%%%%%%%% This paragraph was omitted from the EC submission due to space constraints. 
Note that the algorithm is not obligated to exhaust the budget of $k$ features. When the algorithm chooses not to select any features at all (i.e., $A_t = \emptyset$), the human still needs to make a prediction. In this case, since no additional information is available, the human makes the same prediction of $\hat{y}_h = \bar c$ for every instance of the problem, which represents their belief about the average of the predicted value.

%% file: static.tex
We start by characterizing the algorithm's optimal selection of features in the static case where the human holds fixed beliefs $h$ about feature coefficients. 
In this case, according to \cref{clm:huma-mse}, the algorithm chooses the subset of features minimizing:
% \begin{align*}
$$
MSE(A,h) = (c-\bar c)^2 + \sum_{i \notin A} {a_i}^2 + \sum_{i \in A} (a_i - h_{i})^2 
$$
% \end{align*}
With only one feature, by \cref{clm:huma-mse} we have:
% \begin{align*}
$MSE(\{1\},h) = (c-\bar c)^2 + (a_1 - h_{1})^2$, 
whereas if the algorithm does not select this feature, 
$MSE(\emptyset,h) = (c-\bar c)^2 + {a_1}^2$. 
% \end{align*}
Thus, selecting the feature reduces the error whenever
$a_1^2 > (a_1 - h_1)^2$, which holds if $h_1^2 < 2a_1 h_1$.
For example, when $a_1 \geq 0$, the condition holds  for $h_1 \in (0,2a_1)$.
Note that if the human and algorithm agree on the interpretation of the feature $(a_1 = h_1)$, then this is always satisfied, and so the algorithm would always select the feature. If $a_1$ and $h_1$ have different signs (the human and algorithm completely disagree), then this is never satisfied. 
If they do have the same sign, the feature is useful only if the human's coefficient \( h_1 \) is not too large compared to the algorithm's coefficient \( a_1 \).
When $h_1$ is too large, it means that the human ``overshoots,'' i.e., overuses the feature and ends up too far on the other side of the truth. The factor of $2$ limits overshooting to keep the feature beneficial to the human.

In the single-feature case, we saw that the choice of whether to select the feature or not depends on the value of $MSE(\{1\},h) - MSE(\emptyset,h) = a_1^2-(a_1-h_1)^2$. We find that this is a useful quantity for choosing which features to select in the general case, and refer to it as the {\em value} of feature $1$. As it turns out, this quantity greatly simplifies the problem of computing an optimal subset in the general case. 
More generally, 

\begin{definition} \label{dfn:value-static}
The {\em value} of a subset $A \subseteq [n]$ of features $V(A,h) = MSE(\emptyset,h) - MSE(A,h)$ is the improvement in loss from using the features in $A$ compared to not using any feature. 
\end{definition}

Since for a given problem instance $MSE(\emptyset, h)$ is fixed, to minimize the objective function, the algorithm should select a set $A^*$ of up to $k$ features with the maximum value of $V(A^*,h)$. The following lemma shows that the value of a set of features can be expressed in terms of the values of its individual features.

\begin{lemma} \label{lem:value-static}
The value of a set of features $A\subseteq[n]$ satisfies $V(A,h) = \sum_{i\in A} V(\{i\},h)$.
\end{lemma}
\begin{proof}
\begin{align}
 V(A,h) &= MSE(\emptyset,h) - MSE(A,h)  \notag \\
 &=
(c-\bar c)^2 + \sum_{i=1}^n {a_i}^2  - \big((c-\bar c)^2 + \big(\sum_{i \notin A} {a_i}^2 + \sum_{i \in A} (a_i - h_{i})^2 \big)\big) \notag  \\
&=\sum_{i\in A} a_i^2
- \sum_{i \in A} (a_i - h_i)^2 = \sum_{i\in A} 2a_i h_i-h_i^2  \label{eq:value-static} \\
&= \sum_{i \in A}  
V(\{i\}, h)  \notag
\end{align}
\end{proof}

It is instructive to take a more careful look at 
Equation \eqref{eq:value-static} that the algorithm aims to maximize. 
The contribution of each feature $i \in A$ to the value of a set $A$ is composed of two parts: 
(1) ${a_i}^2$, which captures the informativeness\footnote{The coefficient of a feature captures the feature's importance and variance; recall that features are standardized, and thus coefficients are scaled by the standard deviation, see Lemma \ref{thm:standardized-ftrs} in Appendix \ref{app:general}.} of the feature; (2) minus $(a_i - h_i)^2$, which captures the divergence between the human's and the algorithm's beliefs about the feature. The tension between these two terms is at the heart of this paper.

Lemma \ref{lem:value-static} implies the following corollary for multiple  features. 
\begin{corollary}
    The algorithm only selects features $i$ with a positive value (i.e., $2 a_i h_i > h_i^2 $).  
\end{corollary}

Lemma \ref{lem:value-static} also gives rise to a simple and efficient algorithm to compute an optimal subset of features that the algorithm should select. We first compute the value for each feature $i \in [n]$, as $V(\{i\},h) = 2 a_i  h_i - h_i^2$. Then, we sort features by 
their values and add features to the feature selection set by descending order until reaching the budget of $k$ features or there are no more features with positive values. Thus, we establish that:

\begin{proposition} \label{thm:complexity-static}
An optimal feature selection  $A^* \subseteq [n]$ can be computed in  $n log(n)$ time.
\end{proposition}

%% file: learning1.tex
In this section, we turn to discussing algorithmic assistance for a human decision maker who learns and updates their beliefs through repeated interactions with the algorithm.
In Section \ref{sec:intro}, we saw the fundamental tension between informativeness and divergence, which leads to the phenomenon where the algorithm does not necessarily select 
the most informative features for prediction but instead it may select those features that the human can effectively use. 
As discussed, when the human's beliefs are not fixed but are instead updated through repeated interaction with the algorithm, selecting a less informative set of features -- while may be helpful in the short term before learning occurs -- may limit learning opportunities, and as a result sustain human's incorrect beliefs and potentially lead to worse performance in the long run. We will see that in our framework, this tradeoff is governed by two parameters: $\delta$ that captures how much weight the algorithm's designer puts on short term losses versus long terms losses; and $\phi$ that captures how efficient the human's learning is.

\subsection{The Value of a Sequence }
For the static case (Section \ref{sec:static}), it was useful to define the value of selecting a subset of features (Definition \ref{dfn:value-static}). We start by extending this definition to a sequence of feature subsets in our learning setting.

\begin{definition} \label{dfn:value-learning}
  The {\em value} of a sequence of feature subsets $\seq{S} = (A_t)_{t=0}^\infty$ for a given value of $\delta$ and a learning dynamic that is $\phi$-convergent, is
   the improvement in discounted loss 
    from selecting features according to $\seq{S}$  compared to not using any feature at all. That is,  
\begin{align*}
V_{\delta,\phi}(\seq{S}) &=  L((\emptyset)_{t=0}^\infty,\phi,h_0) -  L(\seq{S},\phi,h_0) 
\end{align*}
\end{definition}
As in the static case, since the loss of not using any feature, $L((\emptyset)_{t=0}^\infty,\phi,h_0)$, is constant, an optimal feature selection is a sequence $\seq{S}^* = (A_t)_{t=0}^\infty$ with maximum value of $V_{\delta,\phi}(\seq{S^*})$ that obeys the budget constraint (i.e., $A_t \subseteq [n]$, $|A_t| \leq k$ for all $t$). The following claim provides an explicit expression for $V_{\delta,\phi}(\seq{S})$. Recall that $\fscore{i}{t}$ is the number of times that feature $i$ was selected prior to time step $t$.
\begin{claim} \label{clm:val-learn-seq}
$V_{\delta,\phi}(\seq{S}=(A_t)_{t=0}^\infty) 
= \sum_{t=0}^\infty \delta^t \sum_{i\in A_t} (a_i^2-\phi(\fscore{i}{t})(a_i-h_{i,0})^2)$
\end{claim}
\begin{proof}
Using the value notion for the non-learning setting (Definition \ref{dfn:value-static}) and \cref{lem:value-static}, we have
\begin{align} \label{eq:value-learning}
V_{\delta,\phi}(\seq{S}) 
& = \sum_{t=0}^\infty \delta^t V(A_t,h_t) \notag 
= \sum_{t=0}^\infty \delta^t \sum_{i\in A_t} V(\{i\},h_{t}) \nonumber \\
&= \sum_{t=0}^\infty \delta^t \sum_{i\in A_t} (a_i^2-\phi(\fscore{i}{t})(a_i-h_{i,0})^2)
\end{align}
\end{proof}

Before we delve into our general results, in the next subsection, 
we illustrate the learning setting in a simple setup with a specific learning dynamic that is $\phi$-convergent. The example demonstrates 
the basic tradeoff between the fixed and growth mindsets (Tradeoff \ref{tradeoff2}) in the algorithm's choice of feature selections. We then continue with the general analysis of the implications of considering a learning decision maker on the algorithm's optimal sequence of feature selections.

\subsection{Example: Exponential Learning Dynamics} \label{sec:example-learning}

For illustration, let us consider
a $\phi$-convergent learning rule given by the following simple dynamic:

\begin{equation} \label{eq:update-rule}
h_{i, t+1} = f(h_{i,t}, a, A_t) =
\begin{cases} 
w \cdot h_{i,t} + (1 - w) \cdot a_i, & \text{if } i \in A_t, \\
h_{i,t}, & \text{otherwise.}
\end{cases}
\end{equation}
where $w \in [0,1]$ represents a learning rate parameter. {Large values of $w$ indicate a human who is a slow learner, while smaller values of $w$ correspond to a faster learner.}
Equivalently, we can write a cumulative version of this step-by-step learning rule, where the human's belief about feature $i$ at time $t$ is given by:
\begin{equation} \label{eq:update-rule-cumulative}
     h_{i,t} = w^{m_i} \cdot h_{i,0}  + (1 - w^{m_i}) \cdot a_i
\end{equation}
and $m_i=\fscore{i}{t}$ denotes the number of times feature $i$ has been selected up to time $t$. At $t=0$, the human starts with their initial beliefs $h_{i,0}$ about each feature $i$. As $m_i \rightarrow \infty$, the human's belief about the coefficient of feature $i$ converges exponentially to the true value $a_i$.

This learning dynamic is $\phi$-convergent with an exponential convergence rate of $\phi(m_i) = w^{2m_i}$. To see why, 
substitute Equation \eqref{eq:update-rule-cumulative} 
and get that:
\begin{align*}
 (a_i-h_{i,t})^2 = (a_i - (w^{m_i} \cdot h_{i,0} + (1-w^{m_i}) a_i))^2 = (w^{m_i}(a_i-h_{i,0}))^2 = w^{2m_i}(a_i-h_{i,0})^2
\end{align*}

 With this learning rule, consider the following toy example. Suppose there are two features and the algorithm can select only one of them at each step. Feature $1$ is more informative than feature $2$, with true coefficients $a_1=1$, $a_2=0.4$.  Initially, the human has a larger error on the more informative feature and a smaller one on the less informative feature, with $h_{1,0} = -0.5$, $h_{2,0} = 0.75$.

 We compare two simple sequences: one in which the algorithm repeatedly selects feature 1 (the ``informative'' sequence) and another in which it repeatedly selects feature 2 (the ``non-divergent'' sequence).  We ask, when does the informative sequence have lower discounted loss than the non-divergent sequence?
 The answer depends on the discount factor $\delta$ and the learning parameter $w$. {When \( \delta \) is very small, the loss is dominated by the first time step, making the non-divergent sequence the better choice, as it provides 
 immediate value. 
 In this case, learning is essentially ignored since its gains occur at a later stage and are heavily discounted.}
 On the other end of the spectrum, when $\delta$ is close to $1$, the algorithm assigns significant weight to later steps where learning plays {an essential} role. In this case, using the informative sequence is preferable, as it provides %\soedit{smaller loss} 
 {greater value} once learning has converged. Thus, we expect a transition point -- an intermediate value of $\delta$ -- above which the informative sequence is preferred and below which the non-divergent sequence is preferred. 
 For higher $w$ values, which indicate slower learning, larger $\delta$ values are required to make the informative sequence preferable.

\begin{figure}[h]
    \vspace{-10pt} % Adjust space before the figure
    \centering
    % Subfigure 1
    \begin{subfigure}[b]{0.49\textwidth}
        \centering
        \includegraphics[width=\textwidth]{"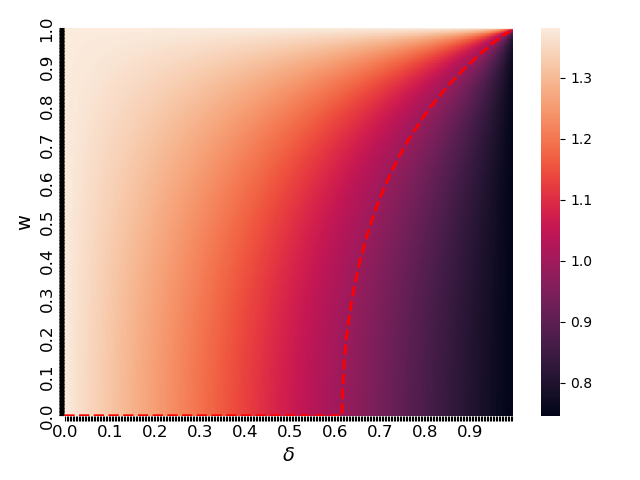"}
        \caption{
        \normalfont A heatmap of the $\delta, w$ plane}
        \label{fig:w-delta-heatmap}
    \end{subfigure}
    \hfill
    % Subfigure 2
    \begin{subfigure}[b]{0.49\textwidth}
        \centering
        \includegraphics[width=1.05\textwidth]{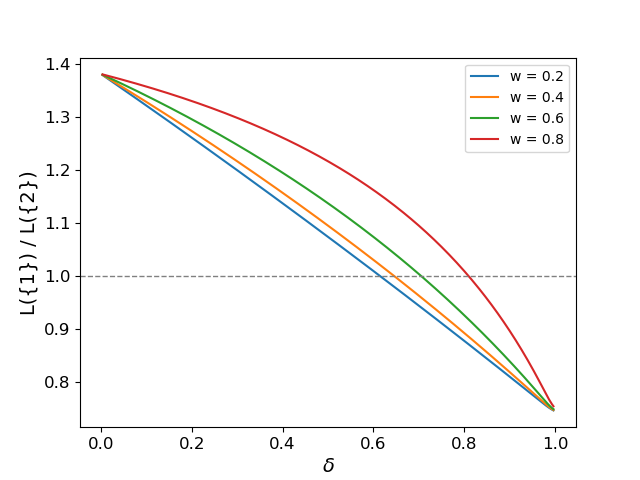}
        \caption{\normalfont Fixing specific values of $w$}
        \label{fig:w-delta-plot}
    \end{subfigure}
    \captionsetup{skip=2pt} % Remove space after caption
    \caption{\normalfont The ratio of discounted losses of the more informative sequence over the less divergent sequence for the example in Section \ref{sec:example-learning}.}
    \label{fig:main_figure}
    \vspace{-10pt} % Adjust space after the figure
\end{figure}

 Figure \ref{fig:w-delta-heatmap} shows a heatmap of the ratio of losses between selecting the informative sequence (feature $1$) and selecting the non-divergent sequence (feature $2$), over the $\delta, w$ plane. The red dashed line indicates a loss ratio of $1$, which is the empirical transition curve; for $\delta$ values to the right of this curve, selecting the informative sequence yields lower loss, while for values to the left of the curve, the non-divergent sequence performs better. 

Specifically, we can see that for every value of $w < 1$ there is a transition point $\delta^*(w)$ above which the algorithm prefers selecting the informative feature. As $w$ increases (slower learning), $\delta^*(w)$ is increasing. The $w=0$ line corresponds to a scenario where the human 
{learns the coefficient of the} feature they observed 
{in one step}. In this example, $\delta^*(w=0) \approx 0.6$. 
The empirical curve in Figure \ref{fig:w-delta-heatmap} 
coincides with our theoretical results characterizing such curves (see Figure \ref{fig:theory-fit-simulation} in Appendix \ref{app:fig-theory-fit-simulation}). Figure \ref{fig:w-delta-plot} shows projections for several $w$ values to further illustrate the transition. 
For slower learning (higher $w$) lines intersect the ratio of 1 (dashed gray line) at higher $\delta$ values.
At the extreme values of $\delta$
 (either close to 0 or 1), we see that the different $w$ values give the same loss ratio; this is because, as explained above, for $\delta$ close to 0, the loss is dominated by the first step, which is independent of $w$, and for $\delta$ close to 1 
the long-term effects of learning dominate, diminishing the influence of finite learning times.

\subsection{Optimality of Stationary Sequences} \label{sec:stationary}

In this section, we analyze optimal sequences of feature selections in the setting of human learning.  
We assume that the magnitudes of the true coefficients are distinct, i.e., $|a_i|\neq |a_j|$ for $i\neq j$. \footnote{Equivalently, $\{a_i^2\}$ has no ties; this can be ensured by an arbitrarily small perturbation that does not affect our conclusions.}

We begin by defining a stationary sequence as a sequence that always selects the same subset of
features.

\begin{definition}
	A sequence of feature subsets $\seq{S} = \infseq{A}$ is {\em stationary} if  $A_t = A \subseteq [n]$ for all $t$. We use the notation $\infsta{A}$ to denote  a stationary sequence that always selects the same subset $A$.
\end{definition}

We are particularly interested in stationary sequences due to their simplicity. We show that for general $n > 1$ and $k \leq n$,  
if the learning dynamic is $\phi$-convergent, then there exists an optimal sequence that is stationary. This fact dramatically decreases the size of the optimization space and the computational complexity of the problem; Later in  \cref{thm:complexity-learning} we show that we can find an optimal stationary sequence in $nlog(n)$ time.

 The proof builds on the following lemma showing that if the learning dynamic is $\phi$-convergent, then any optimal sequence must have a stationary suffix (i.e.,
 an infinite suffix that starting at time step $t$ always selects the same subset).
 
\begin{lemma} \label{lem:constant-suffix}
	If the learning dynamic is $\phi$-convergent, then for any sequence $\seq{S}$ {that does not end in an infinite stationary suffix}, there exists a sequence $\seq{S}'$ of {a strictly} higher value with a stationary suffix.
\end{lemma}

\begin{proof}
Consider a sequence $\seq{S}$ that does not end with an infinite stationary suffix. Denote by \(R(\seq{S})\) the set of features selected infinitely many times. As by definition only the features in \(R(\seq{S})\) are selected infinitely many times, there has to be some time step $T'$ such that for any $t>T'$ only features in \(R(\seq{S})\) are selected. Furthermore, since we consider $\phi$-convergent learning dynamics, there exists time step $T>T'$ after which all features in $R(\seq{S})$ have been selected sufficiently many times for the coefficient of each such feature $i$ to have converged to $a_i$. This implies that their contribution to the accuracy of the prediction is  positive. Hence, if $|R(\seq{S})| \leq k$, the sequence that ends in the stationary suffix that always selects $R(\seq{S})$ has a higher value than the original sequence {(which for $t>T'$ selects subsets of $R(\seq{S})$)}. Similarly, if $|R(\seq{S})| > k$ and there are some steps in which $\seq{S}$ selects less than $k$ features, we can construct a sequence of a higher value by making sure we select $k$ features from $R(\seq{S})$ in each step of the suffix. 
 
We now handle the case that $|R(\seq{S})|>k$ and there exists a suffix in which $\seq{S}$ selects $k$ features at each step. We will show that there exists another sequence $\seq{S}'$ that has a strictly higher value and $|R(\seq{S}')|=|R(\seq{S})|-1$. By applying this argument $|R(\seq{S})|-k$ times we get that there is a sequence $S^*$ with a {strictly} higher value than $\seq{S}$ such that $|R(\seq{S}^*)|=k$. This implies that a sequence with a stationary suffix that always selects $R(\seq{S}^*)$ has a higher value than the value of $\seq{S}$.

Let $A^*$ denote a set that includes the $k$ features in $R(\seq{S})$ that have the highest values of $|a_i|$. By the assumption that the magnitudes of the coefficients are distinct we have that $A^*$ is unique. Note that since the features in $A^*$ have the highest true coefficients, for any feature $i \in A^*$ and $j \in R(\seq{S}) \setminus A^*$, there exists $\eps>0$ such that
	\begin{align} \label{eq-conv-better}
		a_i^2-\eps(a_i - h_{i,0})^2 >a_j^2
	\end{align}
    
	Let $\eps>0$ be such that the inequality above holds for any pair $i \in A^*$ and $j \in R(\seq{S}) \setminus A^*$.  Now, since the learning dynamic is $\phi$-convergent, there exists $m_\eps$ such that for any $m\geq m_\eps$, $\phi(m) \leq \eps$. Let $T_\eps\geq T$ denote the time step by which, for all of the features $i \in A^*$ we have that $m_i \geq m_\eps$. If $|R(\seq{S})|>k$, then there exists some feature $j \in R(\seq{S})\setminus A^*$ that was selected infinitely many times. Since $|A^*|=k$, at each time step $t>T_\eps$ that $j$ was chosen, there exists a feature $i_t \in A^*$ that was not chosen. We create the new sequence $\seq{S}'$, by replacing each selection of feature $j$ in step $t>T_\eps$, by feature $i_t \in A^*$. By construction $|R(\seq{S}')|=|R(\seq{S})|-1$ and by inequality (\ref{eq-conv-better}), the value of the sequence $\seq{S}'$ is greater than the value of sequence $\seq{S}$.
\end{proof}

\begin{theorem} \label{thm:stationary}
	If the learning dynamic is $\phi$-convergent, then there exists an optimal sequence that is stationary (i.e.,  always selects the same subset of features). 
\end{theorem}
\begin{proof}
By \cref{lem:constant-suffix}, we have that 
if an optimal sequence exists, it must end with a stationary suffix. Consider a sequence $\seq{S}=\infseq{A}$ that its stationary suffix begins at $T+1$: $A_T \neq {A_{T+1}}$ and for all $t\geq T+1, A_t = A_{T+1}$. For ease of notation, let $A=A_{T+1}$ and $B=A_{T}$. We will show that there exists a sequence $\seq{S}'$ whose stationary suffix begins at some $T' \leq T$ and attains the same or higher value. Iterating this argument, we obtain that for every sequence there exists a stationary sequence with the same or higher value. Since the number of stationary sequences is finite, it follows that an optimal sequence exists. Hence, an optimal stationary sequence is globally optimal.

We will first compare between three sequences starting from step $T$ (all with the same prefix $A_0,\ldots,A_{T-1}$).

	\begin{enumerate}
		\item $\seq{S}_A$ - a sequence that selects $A$ starting at $T$.
		\item $\seq{S}_{B \rightarrow A} = \seq{S}$ - a sequence that selects $B$ at time $T$ and then selects $A$ for all subsequent time steps. 
		\item $\seq{S}_{B,B\rightarrow A}$ - a sequence that selects $B$ at time steps $T$ and $T+1$ and then selects $A$ for all subsequent time steps.
	\end{enumerate}
% In Appendix \ref{app:3-comp} we prove the following lemma: 
    \begin{lemma} \label{lem:3-comp}
        $V_{\delta,\phi}(\seq{S}_{B \rightarrow A}) < V_{\delta,\phi}(\seq{S}_A)$ or $V_{\delta,\phi}(\seq{S}_{B \rightarrow A}) \leq V_{\delta,\phi}(\seq{S}_{B,B\rightarrow A})$. 
    \end{lemma}
  
    \begin{proof}
    Assume towards contradiction that $V_{\delta,\phi}(\seq S_{B \rightarrow A}) \geq V_{\delta,\phi}(\seq S_A)$ and $V_{\delta,\phi}(\seq S_{B \rightarrow A}) > V_{\delta,\phi}(\seq S_{B,B\rightarrow A})$.
	
	Let $t_i=\fscore{i}{T}$ denote the number of times that feature $i$ was selected prior to time step $T$. We compute the expected discounted values of the sequences we presented, starting at time $T$:
	\begin{align*}
		V_{\delta,\phi}(\seq{S}_A) &= \sum_{t=T}^\infty \delta^t \cdot \sum_{i \in A} (a_i^2 - \phi(t_i+t-T)\cdot(a_i-h_{i,0})^2) \\
		V_{\delta,\phi}(\seq S_{B \rightarrow A}) &= \delta^T \sum_{j \in B}(a_j^2 - \phi(t_j)(a_j-h_{j,0})^2) \\
		&+ \sum_{t=T+1}^\infty \delta^t \cdot \left( \sum_{i \in A \setminus B} (a_i^2 - \phi(t_i+t-T-1)\cdot(a_i-h_{i,0})^2)+ \sum_{i \in A \cap B} (a_i^2 - \phi(t_i+t-T)\cdot(a_i-h_{i,0})^2) \right) \\
		& = \delta^T \sum_{j \in B}(a_j^2 - \phi(t_j)(a_j-h_{j,0})^2)\\
		&+ \sum_{t=T+1}^\infty \delta^t \cdot \left( \sum_{i \in A} a_i^2 -\sum_{i \in A \setminus B} ( \phi(t_i+t-T-1)\cdot(a_i-h_{i,0})^2)-\sum_{i \in A \cap B} ( \phi(t_i+t-T)\cdot(a_i-h_{i,0})^2) \right) \\
		V_{\delta,\phi}(\seq S_{B,B\rightarrow A}) & = \delta^T \sum_{j \in B}(a_j^2 - \phi(t_j)(a_j-h_{j,0})^2)+ \delta^{T+1} \sum_{j \in B}(a_j^2 - \phi(t_j+1)\cdot(a_j-h_{j,0})^2) \\
		&+ \sum_{t=T+2}^\infty \delta^t \cdot \left( \sum_{i \in A} a_i^2 -\sum_{i \in A \setminus B} ( \phi(t_i+t-2-T)\cdot(a_i-h_{i,0})^2)-\sum_{i \in A \cap B} ( \phi(t_i+t-T)\cdot(a_i-h_{i,0})^2) \right) \\
	\end{align*}
	By our assumption $V_{\delta,\phi}(\seq S_{B \rightarrow A}) \geq V_{\delta,\phi}(\seq S_A)$, and hence:
	\begin{align*}
		\delta^T(\sum_{j \in B\setminus A}(&a_j^2 - \phi(t_j)(a_j-h_{j,0})^2)-\sum_{i \in A\setminus B}(a_i^2 - (\phi(t_i))(a_i-h_{i,0})^2)) \\
		&\geq     \sum_{t=T+1}^\infty \delta^t\sum_{i \in A \setminus B} (\phi(t_i+t-1-T)-\phi(t_i+t-T))\cdot(a_i-h_{i,0})^2) 
	\end{align*}
	If we divide by $\delta^T$ and adjust the indices accordingly, we get that:
	\begin{align} \label{eq:BA}
		\sum_{j \in B\setminus A}(&a_j^2 - \phi(t_j)(a_j-h_{j,0})^2)-\sum_{i \in A\setminus B}(a_i^2 - (\phi(t_i))(a_i-h_{i,0})^2) \\
		&\geq   \sum_{t=1}^\infty \delta^t\sum_{i \in A \setminus B} (\phi(t_i+t-1)-\phi(t_i+t))\cdot(a_i-h_{i,0})^2) \notag
	\end{align}
	
	Also by our assumption $V_{\delta,\phi}(\seq S_{B,B\rightarrow A}) < V_{\delta,\phi}(\seq S_{B \rightarrow A})$, we have that:
	\begin{align*} \label{eq:BBA}
		\delta^{T+1} ( \sum_{j \in B\setminus A}(&a_j^2 - \phi(t_j+1)(a_j-h_{j,0})^2)-\sum_{i \in A\setminus B}(a_i^2 - \phi(t_i)(a_i-h_{i,0})^2) \\
		&< \sum_{t=T+2}^\infty \delta^t\sum_{i \in A \setminus B} (\phi(t_i+t-2-T)-\phi(t_i+t-1-T))\cdot(a_i-h_{i,0})^2)
	\end{align*}
	If we divide by $\delta^{T+1}$ and adjust the indices accordingly, we get that:
	\begin{align}
		\sum_{j \in B\setminus A}(&a_j^2 - \phi(t_j+1)(a_j-h_{j,0})^2)-\sum_{i \in A\setminus B}(a_i^2 - \phi(t_i)(a_i-h_{i,0})^2)  \\ &<
		\sum_{t=1}^\infty \delta^{t}\sum_{i \in A \setminus B} (\phi(t_i+t-1)-\phi(t_i+t))\cdot(a_i-h_{i,0})^2)  \notag
	\end{align}

	By combining inequalities (\ref{eq:BA}) and (\ref{eq:BBA}) and some rearranging, we get that:
	\begin{align*}
		\sum_{j \in B\setminus A}(a_j^2 - \phi(t_j)(a_j-h_{j,0})^2) > \sum_{j \in B\setminus A}(a_j^2 - \phi(t_j+1)(a_j-h_{j,0})^2)
	\end{align*}
	By definition $\phi(\cdot)$ is decreasing and hence $(a_j^2 - \phi(t_j+1)(a_j-h_{j,0})^2) \geq (a_j^2 - \phi(t_j)(a_j-h_{j,0})^2)$, in contradiction to the above inequality. 
    \end{proof}

\noindent\textbf{Continuing the proof of Theorem 5.5.} 

Recall from \cref{lem:3-comp} that 
$V_{\delta,\phi}(\seq{S}_{B \rightarrow A}) < V_{\delta,\phi}(\seq{S}_A)$ 
or 
$V_{\delta,\phi}(\seq{S}_{B \rightarrow A}) \leq V_{\delta,\phi}(\seq{S}_{B,B\rightarrow A})$. 
Since $\seq{S}_A$ is a sequence whose infinite suffix starts at $T$ or earlier 
(if $A_{T-1} = A$), if 
$V_{\delta,\phi}(\seq{S}_{B \rightarrow A}) < V_{\delta,\phi}(\seq{S}_A)$, 
then there exists a sequence whose stationary suffix begins at $T' < T{+}1$ 
that attains a higher value, as required. 

Otherwise, $V_{\delta,\phi}(\seq{S}_{B \rightarrow A}) \leq V_{\delta,\phi}(\seq{S}_{B,B\rightarrow A})$. 
Observe that \cref{lem:3-comp} can be applied again to the sequence of $A$’s that begins at $T{+}2$, yielding
\begin{align*}
    V_{\delta,\phi}(\seq{S}_{B,B \rightarrow A}) < V_{\delta,\phi}(\seq{S}_{B \rightarrow A}) 
    \text{~or~} 
    V_{\delta,\phi}(\seq{S}_{B,B \rightarrow A}) \leq V_{\delta,\phi}(\seq{S}_{B,B,B\rightarrow A}). 
\end{align*}
Since $V_{\delta,\phi}(\seq{S}_{B\to A}) \le V_{\delta,\phi}(\seq{S}_{B,B\to A})$, 
it follows that 
$V_{\delta,\phi}(\seq{S}_{B,B\to A}) \le V_{\delta,\phi}(\seq{S}_{B,B,B\to A})$. 
Repeating this argument, we obtain a series of sequences of weakly monotone increasing values with increasing length of the $B$ sequence preceding the stationary suffix, i.e.,
$V_{\delta,\phi}(\seq{S}_{B^{(m)}\to A}) \le
V_{\delta,\phi}(\seq{S}_{B^{(m+1)}\to A})$ for all $m \geq 0$. 
Since the value of any sequence is a discounted sum with geometric decay, 
we have that $\lim_{m\rightarrow \infty} V_{\delta,\phi}(\seq{S}_{B^{(m)}\to A})  = V_{\delta,\phi}(\seq{S}_B)$ and that the convergence is from below, 
where $\seq{S}_{B}$ is a sequence that from step $T$ onward 
selects $B$ and all earlier selections remain identical to $\seq{S}$. 
Since the convergence is from below, we conclude that 
$V_{\delta,\phi}(\seq{S}_B) \geq V_{\delta,\phi}(\seq{S}_{B \rightarrow A})$. 
Thus, we exhibit a sequence with a longer stationary suffix and weakly higher value.

\end{proof}

\subsection{Computing Optimal Stationary Sequences} \label{sec:complexity}
Recall that in Claim \ref{clm:val-learn-seq}, we showed that an optimal sequence for the learning setting maximizes:
\begin{align*}
    V_{\delta,\phi}(\seq{S}=(A_t)_{t=0}^\infty) 
= \sum_{t=0}^\infty \delta^t \sum_{i\in A_t} (a_i^2-\phi(\fscore{i}{t})(a_i-h_{i,0})^2)
\end{align*}

Thus, an optimal stationary sequence chooses a subset $A^* \subseteq[n]$ of at most $k$ features that maximizes:
\begin{align} \label{eq:stationary-i}
V_{\delta,\phi}(\infsta{A^*}) &= \sum_{t=0}^\infty \delta^t \sum_{i\in A^*} (a_i^2-\phi(t)(a_i-h_{i,0})^2) \notag\\
&= \sum_{i\in A^*} \sum_{t=0}^\infty \delta^t (a_i^2-\phi(t)(a_i-h_{i,0})^2) \notag \\
%&= \sum_{i\in A} \left( \frac{1}{1-\delta} a_i^2 - \sum_{t=0}^\infty \delta^t \phi(t)(a_i-h_{i,0})^2 \right) \\
&= \sum_{i\in A^*} \Big(\frac{1}{1-\delta}a_i^2 - \sum_{t=0}^\infty \delta^t \phi(t)(a_i-h_{i,0})^2\Big) \\
&=\sum_{i \in A^*} V_{\delta,\phi}(\infsta{\{i\}}) \notag
\end{align}
Hence, \( A^* \) includes up to $k$ features whose \( V_{\delta,\phi}(\infsta{\{i\}}) > 0 \).  To verify that \( V_{\delta,\phi}(\infsta{\{i\}}) =\frac{1}{1-\delta} a_i^2 - \sum_{t=0}^\infty \delta^t \phi(t)(a_i-h_{i,0})^2 \) is well-defined, observe that \( \sum_{t=0}^\infty \delta^t \phi(t) \) converges. The reason is that since \( \phi(t) \leq 1 \), we have \( \delta^t\phi(t) \leq \delta^t \). Then, by the comparison test, since \( \sum_{t=0}^\infty \delta^t \) converges for $\delta\in (0,1)$, it follows that \( \sum_{t=0}^\infty \delta^t \phi(t) \) also converges. 
In this paper we make the reasonable assumption that $V_{\delta,\phi}(\infsta{\{i\}})$  can be computed in $O(1)$.
As an example, consider the feature values for the exponential learning rule from Section \ref{sec:example-learning}, which is $\phi$-convergent for $\phi(t) = w^{2t}$. 
For this learning rule, 
we obtain $\sum_{t=0}^\infty \delta^t w^{2t} = \frac{1}{1-\delta \cdot w^2}$ and thus the value of feature $i$ is $V_{\delta,\phi}(\infsta{\{i\}}) = \frac{1}{1-\delta} a_i^2 -\frac{1}{1-\delta \cdot w^2} (a_i-h_{i,0})^2$.

As in the static case, this characterization of optimal subsets leads to a simple method for finding an optimal stationary sequence \( S^* \) in \( O(n \log n) \) time, similar to the static case in \cref{thm:complexity-static}: {(1) Compute $ V_{\delta,\phi}(\infsta{\{i\}})$ for each feature $i$, (2) sort them in decreasing order, (3) select the top \( k \) features (or fewer) whose values are positive}. This establishes the following proposition:

\begin{proposition} \label{thm:complexity-learning}
For a given $\delta$ and any $\phi$-convergent learning dynamic, an optimal stationary sequence of feature selections can be computed in $O(n \log(n))$ time.
\end{proposition}

\subsection{When Does the Algorithm Select Informative Features?}
\label{sec:tradeoff-analysis}

Next, we study the following question: under what conditions does the algorithm prioritize selecting the more informative set of features allowing the human to learn (the growth mindset), rather than keep making the myopic choice of the initially less divergent features for short-term rewards (the fixed mindset)? 
As the example in Section \ref{sec:example-learning} illustrates, this depends in our framework on how ``patient'' the algorithm is (the parameter $\delta$) when considering future rewards, as well as the efficiency of human learning (the parameter $\phi$). The definitions and set of results below formalize this intuition. 
\subsubsection{The algorithm's patience}
Recall that $a^2_i$ captures the informativeness of feature $i$, and $(a_i - h_{i, 0})^2$ is the initial divergence of the human's belief about the importance of the feature from the algorithm's estimate. For a pair of features $i$ and $j$ such that $a_i > a_j$, we denote by $\Delta^I_{i,j} = a^2_i - a^2_j$ the \emph{informativeness difference} between features $i$ and $j$, and by $\Delta^D_{i,j} = (a_i - h_{i,0})^2 - (a_j - h_{j,0})^2$ the \emph{divergence difference} between $i$ and $j$. The following lemma uses the informativeness difference and divergence difference between features $i$ and $j$ to characterize the values of $\delta$ for which $V_{\delta,\phi}(\infsta{\{i\}})\geq V_{\delta,\phi}(\infsta{\{j\}})$.
The lemma establishes the main formal basis for analyzing the  \emph{fixed vs. growth mindset} tradeoff (Tradeoff \ref{tradeoff2}). 
It shows that for two features $i$ and $j$, the more informative feature is either always preferred for any $\delta$, or there is a critical value of $\delta$ above which this is true. 
Intuitively, a feature that has high informativeness but also high divergence, may have low value in the early stages  
before learning takes place, but high value in later stages as learning converges. Small $\delta$ puts relatively more weight on the early stages, and so the informative feature might not be selected, but when $\delta$ is large, there is more weight on the outcomes of learning where the more informative feature has higher value.
\begin{lemma} \label{prop:switch-point}
Consider a pair of features $i$ and $j$ such that $|a_i| > |a_j|$. Then, 
\begin{itemize}
    \item If $\Delta^I_{i,j} \geq \Delta^D_{i,j} $, then for any $\delta>0$, $V_{\delta,\phi}(\infsta{\{i\}}) \geq V_{\delta,\phi}(\infsta{\{j\}})$.
    \item  If $\Delta^I_{i,j} < \Delta^D_{i,j}$, there exists a switching point $\delta_{i,j}$ such that, for every $\delta<\delta_{i,j}$, $V_{\delta,\phi}(\infsta{\{i\}}) < V_{\delta,\phi}(\infsta{\{j\}})$ and for every $\delta \geq \delta_{i,j}$, $V_{\delta,\phi}(\infsta{\{i\}}) \geq V_{\delta,\phi}(\infsta{\{j\}})$.
    \end{itemize}
\end{lemma}
\begin{proof}
We first show that there can be at most one value of $\delta$ for which the two features yield equal value, $V_{\delta,\phi}(\infsta{\{i\}}) = V_{\delta,\phi}(\infsta{\{j\}})$. By Equation \eqref{eq:stationary-i}, we should show that the following equation has at most one solution for $\delta_{i,j}$:

\begin{align*}
    \frac{1}{1-\delta_{i,j}} a_i^2 - \sum_{t=0}^\infty \delta_{i,j}^t \phi(t) (a_i-h_{i,0})^2 = \frac{1}{1-\delta_{i,j}} a_j^2 - \sum_{t=0}^\infty \delta_{i,j}^t \phi(t) (a_j-h_{j,0})^2
\end{align*}
By rearranging and 
using our notations $\Delta^I_{i,j}$ and $\Delta^D_{i,j}$, 
\begin{align*}
\frac{1}{1-\delta_{i,j}} \cdot \Delta^I_{i,j} =  \sum_{t=0}^\infty \delta_{i,j}^t \phi(t) \cdot \Delta^D_{i,j} \implies   \frac{\Delta^I_{i,j}}{\Delta^D_{i,j}} = (1-\delta_{i,j})\sum_{t=0}^\infty \delta_{i,j}^t \phi(t)
\end{align*}
Note that 
\begin{align*}
    (1-\delta_i)\sum_{t=0}^\infty \delta_{i,j}^t \phi(t) &= \sum_{t=0}^\infty \delta_{i,j}^t \phi(t) -\sum_{t=0}^\infty \delta_{i,j}^{t+1} \phi(t) \\
    &=1+\sum_{t=1}^\infty \delta_{i,j}^t( \phi(t)-\phi(t-1)) 
    =1-\sum_{t=1}^\infty \delta_{i,j}^t (\phi(t-1)-\phi(t)) 
\end{align*}
Thus, we need to show that there is at most one solution to the equation:
\begin{align} \label{eq:switching-point}
   1- \frac{\Delta^I_{i,j}}{\Delta^D_{i,j}} = \sum_{t=1}^\infty \delta_{i,j}^t (\phi(t-1)-\phi(t))
\end{align}
Note that $\phi(t-1)-\phi(t) \geq 0$ since $\phi$ is decreasing, and since $\phi$ converges to $0$ there exists $t$ such that $\phi(t-1)-\phi(t) > 0$. This means that the sum on the right-hand side of the equation is strictly increasing in $\delta_{i,j}$ and hence Equation \eqref{eq:switching-point} has at most one solution. 

Note that if $\Delta^I_{i,j} \geq \Delta^D_{i,j}$ (i.e., $  a_i^2 - a_j^2 \geq (a_i-h_{i,0})^2 - (a_j-h_{j,0})^2 $ ), the left-hand side is $0$ or negative and hence the equation doesn't have any solution. In this case, for any value of $\delta \in (0,1)$, $V_{\delta,\phi}(\infsta{\{i\}}) \geq V_{\delta,\phi}(\infsta{\{j\}})$. 
If $\Delta^I_{i,j}<\Delta^D_{i,j}$, then the left-hand side is some number between $0$ and $1$. Let $F(\delta) = \sum_{t=1}^\infty \delta^t (\phi(t-1)-\phi(t))$. Note that $F(0) = 0$ and $F(1) = \sum_{t=1}^\infty (\phi(t-1)-\phi(t)) = 1$, where the last equality is because the right-hand side is a telescoping sum that equals $\lim_{t \rightarrow \infty} \phi(0)-\phi(t) = 1$ as $\phi$ converges to $0$. Hence, by the intermediate value theorem Equation \eqref{eq:switching-point} in this case has exactly one solution, thus exactly a single switching point $\delta_{i,j}$, as required. 
\end{proof}

To better understand how the switching point $\delta_{i,j}$ depends on model parameters, it is instructive to consider the closed-form formula of $\delta_{i,j}$ for our running example of a learning dynamic which is $\phi(t) = w^{2t}$-convergent. 

\begin{claim} \label{clm:educating-two-features}
For a $w^{2t}$-convergent learning dynamic, the value of the switching point $\delta_{i,j}$ from \cref{prop:switch-point} is
$$
 \delta_{i,j} 
 = 
 \frac{\Delta^I_{i,j} - \Delta^D_{i,j}}
 {w^2 \cdot \Delta^I_{i,j} - \Delta^D_{i,j}}
 $$
\end{claim}

\begin{proof}
%Recall that $(a_i^2 - a_j^2) = x$ and $\left( (a_i-h_{i,0})^2 - (a_j-h_{j,0})^2 \right) = y$. 
Recall that $\delta_{i,j}$ is the solution of
\begin{align*} 
   1- \frac{\Delta^I_{i,j}}{\Delta^D_{i,j}} = \sum_{t=1}^\infty \delta_{i,j}^t (\phi(t-1)-\phi(t))
\end{align*}
Note that
\begin{align*}
     \sum_{t=1}^\infty \delta_{i,j}^t (\phi(t-1)-\phi(t)) &=  \sum_{t=1}^\infty \delta_{i,j}^t (w^{2(t-1)}-w^{2t}) 
     = \frac{(1-w^2)}{w^2}\sum_{t=1}^\infty \delta_{i,j}^t (w^{2t}) \\
     &= \frac{(1-w^2)}{w^2}\sum_{t=1}^\infty (\delta_{i,j} \cdot w^2)^t 
     = \frac{(1-w^2)}{w^2} \cdot\frac{\delta_{i,j}w^2}{1-\delta_{i,j}w^2} \\
     &= \frac{(1-w^2)\delta_{i,j}}{1-\delta_{i,j}w^2} 
\end{align*}
Now, we look for $\delta_{i,j}$ such that:
\begin{align*} 
\frac{\Delta^D_{i,j}-\Delta^I_{i,j}}{\Delta^D_{i,j}} = \frac{(1-w^2)\delta_{i,j}}{1-\delta_{i,j}w^2} 
\end{align*}
Hence,
\begin{align*} 
&(\Delta^D_{i,j}-\Delta^I_{i,j})\cdot(1-\delta_{i,j}w^2) = \Delta^D_{i,j} \cdot(1-w^2)\delta_{i,j} \\
&\implies \Delta^D_{i,j}-\Delta^I_{i,j} = \delta_{i,j} (\Delta^D_{i,j}-w^2 \Delta^D_{i,j} + w^2 \Delta^D_{i,j}-w^2 \Delta^I_{i,j}) \\
&\implies \delta_{i,j} = 
% \frac{\Delta^D_{i,j}-\Delta^I_{i,j}}{\Delta^D_{i,j}-w^2 \Delta^I_{i,j}}
\frac{\Delta^I_{i,j} - \Delta^D_{i,j}}
{w^2 \cdot \Delta^I_{i,j} - \Delta^D_{i,j}}
\end{align*}
\end{proof}

\cref{prop:switch-point} is the building block of the three results we discuss in this section. We begin by applying it to prove a bound on the number of subsets that are optimal for some value of $\delta \in (0,1)$.

\begin{proposition} \label{prop-delta-bounded}
Fix a problem instance and $\phi$-convergent learning dynamic. Let $A_\phi^*(\delta)$ denote the feature subset selected in an optimal stationary sequence for a patience parameter $\delta$. The number of different subsets for $\delta \in (0,1)$ is at most $\frac{n(n-1)}{2} + k \leq n^2$.
\end{proposition}
\begin{proof}
First, we observe that for small values of \( \delta \), it is possible that there are fewer than \( k \) features with positive values, and hence \( |A_\phi^*(\delta)| \leq k \). Observe that the proof of \cref{prop:switch-point} also implies that for any feature $i$, there is at most a single $\delta$ value for which \( V_{\delta,\phi}(\infsta{\{i\}}) = 0 \). Hence, as we increase $\delta$, there are at most \( k \) switching points, where at each, one additional feature is added to \( A_\phi^*(\delta) \), until we reach a value \( \delta' \) such that for every \( \delta \geq \delta' \), \( |A_\phi^*(\delta)| = k \). 

We can now focus on values of \( \delta \geq \delta' \) for which \( |A_\phi^*(\delta)| = k \). Recall that \( A_\phi^*(\delta) \) consists of the \( k \) features with the highest values. Observe that \( A_\phi^*(\tilde \delta) \neq A_\phi^*(\tilde \delta+\eps) \) for some $\tilde\delta$ and for any \( \eps > 0 \) if, for \(\tilde \delta \), the top \( k \) values correspond to different features than for \(\tilde \delta+\eps \). This implies that \( \tilde \delta \) is a switching point as discussed in \cref{prop:switch-point} of at least two features. \cref{prop:switch-point} also establishes that each pair of features has at most one switching point, implying that the number of switching points and hence optimal subsets of size $k$ is bounded by $\frac{n(n-1)}{2}$. Adding the number of potentially optimal subsets that consist of less then \( k \) features, we obtain an overall bound of at most \( \frac{n(n-1)}{2} + k \) optimal subsets, as required.
\end{proof}

The next theorem describes the impact of $\delta$ on the algorithm's choices. As discussed, when $\delta$ is very small, almost all the value comes from the initial time step before any learning takes place, and feature selection is with a fixed mindset.
We show that as $\delta$ increases and the algorithm puts more weight on future outcomes, it %makes choices with the growth mindset and 
chooses more informative feature subsets (subsets such that  $\sum_{i \in A} a^2_i$ is higher), and so feature selection tends to a growth mindset. We also show that %the last subset in this sequence is the most informative one.
for large enough $\delta$, the most informative subset is selected.

\begin{theorem} \label{prop-delta-efficient} 
    Fix a problem instance and $\phi$-convergent learning dynamic. Let $A_\phi^*(\delta)$ denote the feature subset selected in an optimal stationary sequence for a patience parameter $\delta$. 
    \begin{itemize}
        \item As $\delta$ increases, the informativeness of $A_\phi^*(\delta)$  increases.
        \item There exists  $\delta^* \in (0,1)$ such that for all $\delta > \delta^*$, we have that $A_\phi^*(\delta)$ is the most informative subset of features allowed by the budget $k$. 
    \end{itemize}
\end{theorem}

\begin{proof}
We first prove the statement in the first bullet. As in the proof of \cref{prop-delta-bounded}, we observe that for small values of \( \delta \) the set of optimal features may include less than $k$ features. For such $\delta$ values in each switching point we add more features to the selection set and since $a_i^2$ is positive the informativeness of \( A_\phi^*(\delta) \) can only increase. 
Next, we focus on values of \( \delta \geq \delta' \) for which \( |A_\phi^*(\delta)| = k \). As in the proof of Proposition \ref{prop-delta-bounded}, we observe that if \( A_\phi^*(\tilde \delta) \neq A_\phi^*(\tilde \delta+\eps) \) for some $\tilde\delta$ and for any \( \eps > 0 \), then  \( \tilde \delta \) is a switching point of at least two features.

We now turn to showing that as we increase $\delta$, $A_\phi^*(\delta)$ becomes more informative. Consider the switching point $\delta_{i,j}$, where prior to it, feature $j$ was included in the optimal subset, and after it, feature $j$ is no longer part of the optimal subset, but feature $i$ is. This means that $j$ has a higher value than $i$ for $\delta < \delta_{i,j}$ but a lower value for $\delta > \delta_{i,j}$. By \cref{prop:switch-point}, this implies that $|a_i| > |a_j|$, as required.  

As for the second bulleted statement, let $A^*$ denote the subset of the $k$ most informative features. Consider a feature $i\in A^*$ and $j \not\in A^*$. By \cref{prop:switch-point}, since $a^2_i>a^2_j$, it is either the case that $V_{\delta,\phi}(\infsta{\{i\}}) \geq V_{\delta,\phi}(\infsta{\{j\}})$ for any value of $\delta \in (0,1)$, or there exists $\delta_{i,j} \in (0,1)$ such that for any $\delta \geq \delta_{i,j}$, $V_{\delta,\phi}(\infsta{\{i\}}) \geq V_{\delta,\phi}(\infsta{\{j\}})$. By setting $\delta^*$ to be the maximum of $\delta_{i,j}$'s for any $i\in A^*$ such that $i\in A^*$ and $j \not \in A^*$, and using \cref{thm:stationary}, we get that the optimal sequence is a stationary sequence that selects the most informative subset of features $A^*$.

\end{proof}

\subsubsection{The learning dynamic efficiency} \label{sec-eff-learning}
In this section, we examine the impact of the efficiency of human learning on the algorithm's selection.
We see that, roughly speaking, an optimal subset of features for an efficient learner 
is more informative than that for a
slower learner. 

\begin{definition}
A learning dynamic that is $\phi$-convergent is more efficient than one that is $\phi'$-convergent if $\phi(t) \le \phi'(t)$ for all $t$, and there exists $t$ such that $\phi(t) < \phi'(t)$.
\end{definition}

To take a specific example, in the exponential learning model, a learning dynamic that is 
$w_1^{2t}$-convergent is more efficient than a learning dynamic that is $w_2^{2t}$-convergent if $w_1 < w_2$. At a higher level, we {show} 
that learning dynamics that are $\phi$-convergent and $\phi$ has decreasing marginals (i.e., convex) are more efficient than learning dynamics that are $\phi'$-convergent for $\phi'$ with the same marginals as $\phi$ but in a different order. 
Intuitively, this suggests that it is more beneficial for the human to invest more in learning during earlier time steps rather than later ones, as this allows the algorithm to select more informative features.
Formally, we introduce the following definition to compare the efficiency %speed 
of two learning dynamics.
\begin{definition}
For a $\phi$-convergent learning dynamic, define its marginal function as $\psi(t) = \phi(t-1) - \phi(t)$ for $t \ge 1$.
\end{definition}
By this definition, if $\phi$ has decreasing marginals then $\psi$ is monotonically decreasing. 
\begin{claim}
Consider a learning dynamic that is $\phi'$-convergent, and let $\psi'$ denote its marginal function. 

If $\psi'$ is not decreasing, then the $\phi$-convergent dynamic whose marginals $\psi$ are the same as $\psi'$ but sorted in decreasing order is more efficient.
\end{claim}
\begin{proof}
We need to show that for every $t$, $\phi(t) \leq \phi'(t)$. By definition $\phi(t) = 1-\sum_{x=1}^t \psi(x)$. Thus, we need to show that 
\begin{align*}
 1-\sum_{x=1}^t \psi(x) \leq  1-\sum_{x=1}^t \psi'(x)
\end{align*}
The inequality clearly holds since in the construction of $\phi$ we sorted the values of $\psi$ in decreasing order and hence for any $t$, $\sum_{x=1}^t \psi(x) \geq\sum_{x=1}^t \psi'(x)$, as required.
\end{proof}

The following proposition shows that fixing a problem instance and a $\delta$ value, an optimal sequence for a more efficient learning dynamic selects the same or more informative feature subset. 

\begin{proposition} \label{prop-phi-increases}
    Let $A^{*}_{\delta}(\phi)$ denote the subset of features chosen in an optimal stationary sequence. If $\phi$ is more efficient than $\phi'$, then $A^{*}_{\delta}(\phi)$ is more informative than $A^{*}_{\delta}(\phi')$.
\end{proposition}
\begin{proof}
    Recall that an optimal subset includes $k$ features with highest values $V_{\delta,\phi}(\infsta{\{i\}}) = \sum_{t=0}^\infty \delta^t (a_i^2-\phi(t)(a_i-h_{i,0})^2)$ (given that those are positive). 
    We show that if $|a_i|>|a_j|$, then:
    \begin{itemize}
\item $V_{\delta,\phi}(\infsta{\{i\}}) \leq V_{\delta,\phi}(\infsta{\{j\}}) \implies V_{\delta,\phi'}(\infsta{\{i\}}) \leq V_{\delta,\phi'}(\infsta{\{j\}})$. 
\item $V_{\delta,\phi'}(\infsta{\{i\}}) \geq V_{\delta,\phi'}(\infsta{\{j\}}) \implies
V_{\delta,\phi}(\infsta{\{i\}}) \geq V_{\delta,\phi}(\infsta{\{j\}})$.
    \end{itemize}
    This means that for every pair of features, either both agree on which feature has the higher value, or only learning dynamics that are $\phi$-convergent assign feature $i$ a higher value than feature $j$. As a result, the set of optimal features selected by a $\phi$-convergent learning dynamic is at least as informative as the set of features selected by a $\phi'$-convergent learning dynamic.

    Recall that the informativeness difference between features $i$ and $j$ is  $\Delta^I_{i,j} = a^2_i - a^2_j$ and the divergence difference is $\Delta^D_{i,j} = (a_i - h_{i,0})^2 - (a_j - h_{j,0})^2$. Observe that:
    \begin{align*}
    V_{\delta,\phi}(\infsta{\{i\}}) - V_{\delta,\phi}(\infsta{\{j\}}) &= \sum_{t=0}^\infty \delta^t (a_i^2-\phi(t)(a_i-h_{i,0})^2) - \sum_{t=0}^\infty \delta^t (a_j^2-\phi(t)(a_j-h_{j,0})^2)  \\
    &= \frac{1}{1-\delta} \Delta^I_{i,j} - \sum_{t=0}^\infty \delta^t \cdot \phi(t) \cdot \Delta^D_{i,j}
    \end{align*}
    Since $|a_i|>|a_j|$ we have that $\Delta^I_{i,j}>0$. If $\Delta^D_{i,j} \leq 0$, then for any learning dynamic that is $\tilde \phi$-convergent, we have that $V_{\delta,\tilde\phi}(\infsta{\{i\}}) - V_{\delta,\tilde\phi}(\infsta{\{j\}}) > 0$. The more interesting case is when $\Delta^D_{i,j} > 0$, i.e., feature $i$ is more divergent than feature $j$. For this case we observe that 
    $
\sum_{t=0}^\infty \delta^t \cdot \phi(t) \leq \sum_{t=0}^\infty \delta^t \cdot \phi'(t) 
    $,
and hence 
$
V_{\delta,\phi}(\infsta{\{i\}}) - V_{\delta,\phi}(\infsta{\{j\}}) \geq V_{\delta,\phi'}(\infsta{\{i\}}) - V_{\delta,\phi'}(\infsta{\{j\}})
$
which implies that both bulleted statements hold, as required.   
\end{proof}

%% file: misspecification-full-text.tex
In this section, we analyze the effect of errors in the algorithm's estimates of the ground-truth coefficients $a$, the human's coefficients $h$, and the convergence rate $\phi$ of the learning dynamic. Modeling errors can lead to incorrect feature selection, reducing overall value. To quantify this, we express these errors as the maximum possible error margin in value per feature. As we will see, it is important to distinguish between "overshoot" errors and ``undershoot'' errors, as their error margins can be asymmetric. The following definition captures this formally. For the rest of this section $V(\{i\})$ is the
true value of feature $i$ depending on whether we are in the static setting or the learning setting, and $V'(\{i\})$ is the value computed by the imperfect algorithm.

\begin{definition} \label{def:error-margins}
Let $\eps_i$ denote an upper bound on the magnitude of  error in some coefficient of feature $i$ (e.g., ground-truth coefficient or human's coefficient). The error margin of feature $i$'s value, $V(\{i\})$, is defined by two non-negative functions $\bar \xi_i(\eps_i)$ and $\xi_i(\eps_i)$, such that,
\begin{align*}
    V(\{i\})-\bar \xi_i(\eps_i) \leq V'(\{i\})  \leq  V(\{i\}) +\xi_i(\eps_i)
\end{align*}
\end{definition}

The error margins in estimating specific values can be used to compute the error margin in the algorithm's selection. The idea is to incorporate these margins into the algorithm's decision of whether to include feature $i$ or feature $j$. Specifically, let $A^*$ denote an optimal subset. If for feature $i \in A^*$ and any feature $j \not \in A^*$, we have that $V(\{i\}) - V(\{j\}) \geq \bar{\xi}_i(\epsilon_i) + \xi_j(\epsilon_j)$, then the imperfect algorithm will also prefer $i$ over $j$, and the error will not affect the result. Else, the algorithm may choose $j$ instead of $i$, but the error from this choice will be bounded by $\xi_i(\epsilon_i) + \xi_j(\epsilon_j)$. 

Note that we do not assume anything about the structure of the error (e.g., the direction of the error or on which features the algorithm errs). 
Since the actual performance is affected only by the ranking of features by their values, increasing $\eps$ can have impact only on a collection of thresholds where ranking changes. That is, the change in value is discontinuous and in particular, it is a composition of step functions. 
\begin{proposition}\label{prop-mis-gen-error}
 Let $A^*$ denote an optimal feature set and $A$ denote the feature set selected by a misspecified algorithm.  Let $\eps_i$ be a bound on the magnitude of the estimation error for some coefficient of feature $i$, then:
$% \begin{align*}
     V(A^*)-V(A)  \leq  \sum_{i\in  A* \setminus A} \bar \xi_i(\eps_i) +\sum_{j\in  A \setminus A^*} \xi_j(\eps_j).
 $ %\end{align*}
\end{proposition}
\begin{proof}
Note that, 
\begin{align*}
    V(A^*)-V(A) = \sum_{i\in A^* \setminus A} V(\{i\}) - \sum_{j\in A \setminus A^*} V(\{j\}) = 
   {\sum_{\text{unique} (i,j), i\in  A^* \setminus A, j\in  A \setminus A^*}}
    V(\{i\}) - V(\{j\})
\end{align*}
Where the last step is a partition of the features in $A^* \setminus A$ and $A \setminus A^*$ to distinct pairs $i$ and $j$. If $|A^*|\neq|A|$ we can introduce some dummy features that have value $0$ (i.e., $a=a'=h=h'=0)$ to make the two sets equal without affecting anything else. For every pair $i \in A^* \setminus A$ and $ j \in A \setminus A^*$, we get
\begin{align*}
    V'(\{i\})-V'(\{j\}) \geq V\{i\} - \bar \xi_i(\eps_i) - (V(\{j\}) + \xi_j(\eps_j) ) = V(\{i\}) - V(\{j\}) - (\bar \xi_i(\eps_i) + \xi_j(\eps_j)) 
\end{align*}
Thus, if the difference in values is large, $V(\{i\}) - V(\{j\}) \geq \bar \xi_i(\eps_i) + \xi_j(\eps_j)$, the misspecified algorithm should also select feature $i$ instead of feature $j$. Since it did not, we know that the difference between the features is small and hence, $V(\{i\}) - V(\{j\}) \leq \bar \xi_i(\eps_i) + \xi_j(\eps_j)$, which completes the proof.
\end{proof}

Recall that the value of a feature set~$A$ is defined as
$V(A, h) = \mathrm{MSE}(\emptyset, h) - \mathrm{MSE}(A, h)$.
It is straightforward to see from this definition that
$V(A^*) - V(A) = \mathrm{MSE}(A, h) - \mathrm{MSE}(A^*, h)$.
Therefore, the error margins in value are identical to the corresponding
error margins in the mean--squared loss, which is ultimately the
quantity of interest.

In the rest of this section we go over the possible misspecifications of the algorithm of the ground-truth coefficients and human's coefficients in the static setting (S) and in the learning setting (L) and compute functions $\bar \xi_i(\eps_i),  \xi_i(\eps_i)$ as in Definition \ref{def:error-margins}. Table \ref{tab:xi_margins} summarizes the values of $\bar \xi_i(\eps_i)$. The value of $\xi_i(\eps_i)$ is the same except that the sign of the $\eps^2$ term (if exists) is flipped:  

\begin{table}[h]
    \centering
    \begin{tabular}{|c|c|c|}
        \hline
        Error & Requirements & $\bar \xi_i(\eps_i)$ \\
        \hline
        S: $|a'_i - a_i| \leq \eps_i$ & & $2\eps_i |h_i|$ \\
        \hline
        S: $|h'_i - h_i| \leq \eps_i$ & $|h_i - a_i| \geq \eps_i$ & $2\eps_i |h_i - a_i| + \eps_i^2$ \\
        \hline
        L: $|h'_i - h_i| \leq \eps_i$ & $|h_i - a_i| \geq \eps_i$ & $\sum_{t=0}^\infty \delta^t \phi(t)(2\eps_i |a_i - h_{i,0}| + \eps_i^2)$ \\
        \hline
        L: $|a'_i - a_i| \leq \eps_i$ & $|a_i| \geq \eps_i, |a_i-h_i| \geq \eps_i$ & $ 
        2\eps_i \left|\frac{1}{1-\delta} a_i - \sum_{t=0}^\infty \delta^t \phi(t)(a_i - h_{i,0})\right| - \eps_i^2(\frac{1}{1-\delta} - \sum_{t=0}^\infty \delta^t \phi(t))$ \\
        \hline
    \end{tabular}
    \caption{\normalfont The value of $\bar \xi_i(\eps_i)$ for different types of misspecifications.}
    \label{tab:xi_margins}
      \vspace{-10pt} % Adjust space after the figure
\end{table}

Let $A$ denote the feature set chosen by an imperfect algorithm and $A^*$ an optimal set {according to the true parameters}. We will show that the functions $\bar \xi_i(\eps_i), \xi_i(\eps_i)$ are increasing in $\eps_i$ (for some types of error as long as $\eps_i$ below a required threshold). Thus, we can apply \cref{prop-mis-gen-error} to get an upper bound on the total error of the algorithm. Note that in all the rows of the table, except for the first one, the error is for $|A|=|A^*|$ and $\oplus$ denotes the symmetric difference between two sets. If $|A|\neq |A^*|$ there will be additional terms as specified by \cref{prop-mis-gen-error}. Our results are summarized in Table \ref{table-total-error}.

\begin{table}[h]
    \centering
    \begin{tabular}{|c|c|c|}
        \hline
        Error & Requirements & $V(A^*) - V(A) \leq$ \\
        \hline
        S: $\forall i, |a'_i - a_i| \leq \eps$ & & $2\eps \sum_{i \in A^* \oplus A} |h_i|$ \\
        \hline
        S: $\forall i,|h'_i - h_i| \leq \eps$ & $|h_i - a_i|\geq \eps $ & $2\eps \sum_{i \in A^* \oplus A} |h_i - a_i|$ \\
        \hline
        L: $\forall i,|h'_i - h_i| \leq \eps$ & $ |h_i - a_i|\geq \eps $ & $2\eps \sum_{t=0}^\infty \delta^t  \sum_{i \in A^* \oplus A} |h_i - a_i|$ \\
        \hline
        L: $\forall i,|a'_i - a_i| \leq \eps$ & $ |a_i|\geq \eps,  |a_i-h_i| > \eps$ & $2\eps \sum_{i \in A^* \oplus A } \left|\frac{1}{1-\delta} a_i - \sum_{t=0}^\infty \delta^t \phi(t)(a_i - h_{i,0})\right|$ \\
        \hline
    \end{tabular}
    \caption{\normalfont Error margins for different types of misspecifications.}
    \label{table-total-error}
  \vspace{-10pt} % Adjust space after the figure
\end{table}

For the learning setting when $\phi$ is misspecified such that  $|\sum_{t=0}^\infty \delta^t \phi'(t) - \sum_{t=0}^\infty \delta^t \phi(t)| \leq \eps$ and the algorithm is correct about all coefficients, we show that, 
$V_{\delta,\phi}(\infsta{A^*})-V_{\delta,\phi}(\infsta{A})) \leq  \eps \sum_{i \in A^* \cup A \setminus A^*\cap A } (a_i-h_{i,0})^2.$ 

In the remainder of this section we will consider the different misspecifications we discussed and prove bounds on $\bar \xi_i(\eps_i)$ and $\xi_i(\eps_i)$, by applying Proposition \ref{prop-mis-gen-error} we get the error detailed in Table \ref{table-total-error}.

\subsection{Misspecification in the Non-Learning Setting}
Recall that when the human's beliefs are fixed, we have $V(\{i\},h) = 2a_ih_i-h_i^2$.
We first bound the error resulting from the algorithm misspecifying the ground-truth coefficients. As we will see, in this case the value's margin error is symmetric.

\begin{observation}
Consider a fixed human belief setting and feature $i$ such that $|a'_i-a_i|\leq \eps_i$ and $h'_i= h_i$ for every $i$, then $\bar \xi_i(\eps_i)=\xi_i(\eps_i) =  2\eps_i |h_i|$.
\end{observation}
\begin{proof}
Let $\eps' = |a'_i-a_i|$. Observe that,
    \begin{align*}
        V'(\{i\}) = 2a'_ih_i-h_i^2 = 2(a_i\pm \eps')h_i-h_i^2 = V(\{i\}) \pm 2\eps' h_i
    \end{align*}
    Hence, we get that: $V'(\{i\}) \geq V(\{i\}) - 2\eps'|h_i|$ and $V'(\{i\}) \leq V(\{i\}) + 2\eps'|h_i|$. Since both errors are increasing in $\eps'$ in their respective directions, we conclude that $\bar \xi_i(\eps_i)=\xi_i(\eps_i) =  2\eps_i |h_i|$ as required.
\end{proof}
By applying \cref{prop-mis-gen-error} for all features $i$ and assuming $\eps_i=\eps$, we get:
\begin{corollary}
In the non-learning setting, if for every $i$, $|a'_i-a_i|\leq \eps$ and $h'_i=h$, then $V(A^*)-V(A) \leq 2\eps \sum_{i \in A^* \oplus A} |h_i| $
\end{corollary}

Next, we turn to bound the error resulting from the algorithm misspecifying the human's coefficients.
\begin{observation} \label{obs-err-h-static}
Consider a fixed human belief setting and feature $i$ such that $|h'_i-h_i|\leq \eps_i$, $a'_i= a_i$, and $\eps \leq |h_i-a_i|$, then $\bar \xi_i(\eps_i)=2\eps_i|h_i-a_i|+(\eps_i)^2$ and $\xi_i(\eps_i)=2\eps_i|h_i-a_i|-(\eps_i)^2$.
\end{observation}
\begin{proof}
Let $\eps' = |h'_i-h_i|$. Observe that,
\begin{align*}
V'(\{i\}) &= 2a_ih'_i-(h'_i)^2 = 2a_i(h_i \pm \eps')-(h_i\pm \eps')^2 = V(\{i\}) \pm 2a_i\eps' \mp 2\eps' h_i - (\eps')^2 \\
&= V(\{i\}) \pm 2\eps'|h_i-a_i|-(\eps')^2
\end{align*}
 We get that, $V'(\{i\}) \geq V(\{i\}) -(2\eps'|h_i-a_i|+(\eps')^2)$ as this error margin increases with $\eps'$, we get that $\bar \xi_i(\eps_i)=2\eps_i|h_i-a_i|+(\eps_i)^2$. Now, observe that, $V'(\{i\}) \leq V(\{i\}) + \max\{2\eps'|h_i-a_i|-(\eps')^2,0\}$.  If $h_i$ and $a_i$ are extremely close to one another we have an undershoot error margin of roughly $\eps'^2$. Hence, the larger error occurs when $|h_i-a_i|\geq \eps'$. This is also the more plausible case as it assumes that the error of the algorithm is smaller than the error of the human. Hence, we conclude that $\xi_i(\eps_i)=2\eps_i|h_i-a_i|-(\eps_i)^2$.
\end{proof}
By applying \cref{prop-mis-gen-error} for all features $i$ and assuming $\eps_i=\eps$, we get:
\begin{corollary}
In the non-learning setting, if for every feature $i$, $|h'_i-h_i|\leq \eps$, $|h_i-a_i|\geq \eps$ and $a'_i= a_i$, then for $|A|=|A^*|$ we have that $V(A^*)-V(A) \leq   2\eps \sum_{i \in A^* \oplus A} |h_i-a_i|$. If $|A|\neq|A^*|$, the bound includes additional terms as specified by \cref{prop-mis-gen-error}.
\end{corollary}

It is interesting to note that when the human agrees with the algorithm on the sign of the coefficients, the error for misspecifying the ground-truth coefficient can be substantially larger than the error for misspecifying the human coefficients. The situation is reversed when the coefficients of the human and the ground truth have  opposite signs. 

\subsection{Misspecification in the Learning Setting}
Recall that in this setting:
 \begin{align*} 
 V_{\delta,\phi}(\infsta{\{i\}}) =   \frac{1}{1-\delta}a_i^2 - \sum_{t=0}^\infty \delta^t \phi(t)(a_i-h_{i,0})^2
 \end{align*}

 We first consider an algorithm that misspecifies the human's coefficients and observe that the error margins are {direct generalization} of the error we have seen for the analogous error in the non-learning setting.
 \begin{observation} \label{obs-err-h-learning}
Consider the learning setting such that $\phi'=\phi$, and some feature $i$ such that $|h'_i-h_i|\leq \eps_i$, $a'_i= a_i$, and $|h_i-a_i|\geq \eps$, then $\bar \xi_i(\eps_i)= \sum_{t=0}^\infty \delta^t \phi(t)\big(2\eps_i \cdot (|a_i-h_{i,0}|) +\eps_i^2 \big)$ and 
$\xi_i(\eps_i)  = \sum_{t=0}^\infty \delta^t \phi(t)\big(2\eps_i \cdot (|a_i-h_{i,0}|) -\eps_i^2 \big)$.
\end{observation}
\begin{proof}
    Let $\eps'=|h'_i-h_i|$. Observe that,
    \begin{align*}
   V'_{\delta,\phi}(\infsta{\{i\}}) &=   \frac{1}{1-\delta}a_i^2 - \sum_{t=0}^\infty \delta^t \phi(t)(a_i-h'_{i,0})^2 \\
   &= \frac{1}{1-\delta}a_i^2 - \sum_{t=0}^\infty \delta^t \phi(t)(a_i-h_{i,0}\pm \eps')^2 \\
   &= \frac{1}{1-\delta}a_i^2 - \sum_{t=0}^\infty \delta^t \phi(t)(a_i-h_{i,0})^2 -  \sum_{t=0}^\infty \delta^t \phi(t)\big((\eps')^2 \pm  2\eps' \cdot (a_i-h_{i,0}) \big) \\
   &= V_{\delta,\phi}(\infsta{\{i\}}) -  \sum_{t=0}^\infty \delta^t \phi(t)\big((\eps')^2 \pm  2\eps' \cdot (a_i-h_{i,0}) \big)
\end{align*}
We skip the rest of the analysis as it is the same as in \cref{obs-err-h-static}.

\end{proof}
By applying \cref{prop-mis-gen-error} for all features $i$ and assuming $\eps_i=\eps$, we get:
\begin{corollary}
If $\phi'=\phi$ and for every feature $i$, $|h'_i-h_i|\leq \eps$,  $|h_i-a_i| \geq \eps$ and $a'_i= a_i$, then for $|A|=|A^*|$ we have that,
\begin{align*}
    V_{\delta,\phi}(\infsta{A^*})-V_{\delta,\phi}(\infsta{A})) \leq    2\eps\sum_{t=0}^\infty \delta^t \sum_{i \in A^* \oplus A} |h_i-a_i|
\end{align*}
\end{corollary}

Next, we consider an imperfect algorithm with inaccurate ground-truth coefficients. In the learning setting, the effect of such errors is more nuanced than in the static setting. First, they {capture} the informativeness of the {features}. Second, even when the algorithm initially has a correct estimate of the humans' coefficients, as the human learns based on the ground truth, the algorithm's estimate of the human's coefficients becomes inaccurate. As we will see, this leads to an interesting interplay between these two types of errors.

Our analysis can be viewed as an upper bound on the algorithm's error since, in practice, the algorithm is also likely to learn and improve its estimate of the ground-truth coefficients over time. The interaction in this case is significantly more complex, and we defer its detailed analysis to future work.

\begin{observation} \label{obse-err-a-learning}
Consider the learning setting such that $\phi'=\phi$, and some feature $i$ such that $|a'_i-a_i|\leq \eps_i$, $h'_i= h_i$ and $ \eps_i \leq \min\{|a_i|,|a_i-h_i|\}$, then,
\begin{itemize}
\item  $\xi_i(\eps_i) =   2\eps_i \big|\frac{1}{1-\delta} a_i - \sum_{t=0}^\infty \delta^t \phi(t)(a_i-h_{i,0})\big| + \eps_i^2\big(\frac{1}{1-\delta}-\sum_{t=0}^\infty \delta^t \phi(t)\big)$.
\item $\bar \xi_i(\eps_i) = 2\eps_i \big|\frac{1}{1-\delta} a_i - \sum_{t=0}^\infty \delta^t \phi(t)(a_i-h_{i,0})\big| - \eps_i^2\big(\frac{1}{1-\delta} -\sum_{t=0}^\infty \delta^t \phi(t) \big)$.
\end{itemize}
\end{observation}
\begin{proof}
     Let $\eps'=|a'_i-a_i|$. Observe that,
\begin{align*}
   V'_{\delta,\phi}(\infsta{\{i\}}) &=   \frac{1}{1-\delta}(a'_i)^2 - \sum_{t=0}^\infty \delta^t \phi(t)(a'_i-h_{i,0})^2 \\
   &= \frac{1}{1-\delta}(a_i\pm \eps')^2 - \sum_{t=0}^\infty \delta^t \phi(t)(a_i-h_{i,0}\pm \eps')^2 \\
   &= \frac{1}{1-\delta}a_i^2 - \sum_{t=0}^\infty \delta^t \phi(t)(a_i-h_{i,0})^2 + \frac{1}{1-\delta}((\eps')^2 \pm 2a_i \eps') -  \sum_{t=0}^\infty \delta^t \phi(t)\big((\eps')^2 \pm  2\eps' \cdot (a_i-h_{i,0}) \big) \\
   &=V_{\delta,\phi}(\infsta{\{i\}})  + \frac{1}{1-\delta}(\eps_i^2 \pm 2a_i \eps') -  \sum_{t=0}^\infty \delta^t \phi(t)\big(\eps'^2 \pm  2\eps' \cdot (a_i-h_{i,0}) \big)\\
   &= V_{\delta,\phi}(\infsta{\{i\}})+ (\eps')^2\Big(\frac{1}{1-\delta}-\sum_{t=0}^\infty \delta^t \phi(t)\Big) \pm \frac{1}{1-\delta} (2a_i\eps')\mp \sum_{t=0}^\infty \delta^t \phi(t)2\eps'(a_i-h_{i.0})
\end{align*}
We get that:
\begin{align*}
    V'_{\delta,\phi}(\infsta{\{i\}}) &\geq V_{\delta,\phi}(\infsta{\{i\}})-\bigg( \bigg|\frac{1}{1-\delta} (2a_i\eps') - \sum_{t=0}^\infty \delta^t \phi(t)2\eps'(a_i-h_{i,0})\bigg| - \eps'^2\big(\frac{1}{1-\delta}-\sum_{t=0}^\infty \delta^t \phi(t)\big) \bigg) \\
 V'_{\delta,\phi}(\infsta{\{i\}}) &\leq V_{\delta,\phi}(\infsta{\{i\}})+ \bigg( \bigg|\frac{1}{1-\delta} (2a_i\eps') - \sum_{t=0}^\infty \delta^t \phi(t)2\eps'(a_i-h_{i,0})\bigg|+\eps'^2\big(\frac{1}{1-\delta}-\sum_{t=0}^\infty \delta^t \phi(t)\big) \bigg)
\end{align*}

Note that the second term of the lower bound is positive since 
 $\frac{1}{1-\delta} \geq \sum_{t=0}^\infty \delta^t \phi(t)$. Hence, a-priori it is unclear that the error is increasing with $\eps'$. By taking a derivative with respect to $\eps'$ and checking when is it positive, we get

$$ 2\bigg|\frac{1}{1-\delta} a_i - \sum_{t=0}^\infty \delta^t \phi(t)(a_i-h_{i,0})\bigg| - 2 \eps' ( \frac{1}{1-\delta} -  \sum_{t=0}^\infty \delta^t \phi(t)) \geq 0 $$
Implying the following condition on $\eps'$:
\begin{align*}
    \eps' \leq \frac{\big|\frac{1}{1-\delta} a_i - \sum_{t=0}^\infty \delta^t \phi(t)(a_i-h_{i,0})\big|}{\frac{1}{1-\delta} -\sum_{t=0}^\infty \delta^t \phi(t)} 
\end{align*}
Observe that this condition holds as long as $\eps_i \leq \min\{|a_i|,|a_i-h_i|\}$
 Hence, we conclude that,
 \begin{align*}
\bar \xi_i(\eps_i) = 2\eps' \big|\frac{1}{1-\delta} a_i - \sum_{t=0}^\infty \delta^t \phi(t)(a_i-h_{i,0})\big| -\eps'^2\big(\frac{1}{1-\delta}-\sum_{t=0}^\infty \delta^t \phi(t)\big)
 \end{align*}
We get that the upper bound is increasing in $\eps'$ since $\frac{1}{1-\delta}-\sum_{t=0}^\infty \delta^t \phi(t) \geq 0$ and thus,
\begin{align*}
\xi_i(\eps_i) =  2\eps' \big|\frac{1}{1-\delta} a_i - \sum_{t=0}^\infty \delta^t \phi(t)(a_i-h_{i,0})\big| + \eps'^2\big(\frac{1}{1-\delta}-\sum_{t=0}^\infty \delta^t \phi(t)\big)
\end{align*}
\end{proof}
By applying \cref{prop-mis-gen-error} for all features $i$ and assuming $\eps_i=\eps$, we get:
\begin{corollary}
If $\phi'=\phi$ and for every feature $i$, $|a'_i-a_i|\leq \eps$, $  \min\{|a_i|,|a_i-h_i| \geq \eps$ and $h'_i= h_i$, then for $|A|=|A^*|$ we have that
    \begin{align*}
        V(A^*)-V(A) \leq   2\eps \sum_{i \in A^* \oplus A} \big|\frac{1}{1-\delta} a_i - \sum_{t=0}^\infty \delta^t \phi(t)(a_i-h_{i,0})\big|
    \end{align*}
\end{corollary}
It is interesting to further examine the error estimate we obtained. Note that if $a_i$ and $a_i - h_i$ have the same sign, the two types of errors partially cancel out, reducing the total error. However, if they have opposite signs, they compound, increasing the total error.

Lastly, we look at errors in estimating the 
human's learning dynamic $\phi$. There are various measures that can be used to specify a bound on the distance between the true function $\phi$ and the estimated function $\phi'$. In our case, since $\phi$ always appears in converging sums, it is useful to define the error $\eps$ in estimating %the convergence rate, 
$\phi$, given the patience parameter $\delta$, as the error in estimating the sum $\sum_{t=0}^\infty \delta^t \phi(t)$. 
As explained in Section \ref{sec:model}, this quantity summarizes the weight of the human's initial divergence $(a_i - h_{i,0})^2$ in a feature's value estimate; when learning is fast, the weight is small, and when learning is slow, the weight is large. The error in value of a feature, given such an error in $\phi$, is: 

\begin{observation}
Consider an algorithm that assumes that human learning is $\phi'$-convergent, where in fact it is $\phi$-convergent, such that,  $|\sum_{t=0}^\infty \delta^t \phi'(t) - \sum_{t=0}^\infty \delta^t \phi(t)| \leq \eps$ and some feature $i$ such that $h'_i=h_i$ and $a'_i= a_i$, then 
{$\bar \xi_i(\eps) = \xi_i(\eps) = \eps (a_i-h_{i,0})^2$}.
% $\bar \xi_i(\eps)= ?$ and $\xi_i(\eps)$.
\end{observation}
\begin{proof}
Let $\eps' = |\sum_{t=0}^\infty \delta^t \phi'(t) - \sum_{t=0}^\infty \delta^t \phi(t)|$
\begin{align*}
   V_{\delta,\phi'}(\infsta{\{i\}}) &=   \frac{1}{1-\delta}(a_i)^2 - \sum_{t=0}^\infty \delta^t \phi'(t)(a_i-h_{i,0})^2 \\
   &= \frac{1}{1-\delta}(a_i)^2 - \big(\sum_{t=0}^\infty \delta^t \phi(t) \pm\eps \big)(a_i-h_{i,0})^2 \\ 
   &=   V_{\delta,\phi}(\infsta{\{i\}}) \pm \eps (a_i-h_{i,0})^2
\end{align*}
 As it is easy to see that the marginal error is increasing in $\eps'$, we conclude that, $\bar \xi_i(\eps) = \xi_i(\eps) = \eps (a_i-h_{i,0})^2$.   
\end{proof}

By applying \cref{prop-mis-gen-error} for all features $i$ and assuming $\eps_i=\eps$, we get:
 \begin{corollary}
  If $|\sum_{t=0}^\infty \delta^t \phi'(t) - \sum_{t=0}^\infty \delta^t \phi(t)| \leq \eps$ and for every feature $i$, $h'_i=h_i$ and $a'_i= a_i$, then, 
$V_{\delta,\phi}(\infsta{A^*})-V_{\delta,\phi}(\infsta{A})) \leq  \eps \sum_{A^* \oplus A } (a_i-h_{i,0})^2.$

 \end{corollary}
 
So we see that the impact of misestimating the speed of learning on the value estimate is the product of two factors: 
the extent of the error in $\phi$ and the magnitude of the human's initial error in their belief.
 If the human's initial belief about a feature is already close to its true coefficient, misestimating the speed of learning does not matter much since there is little for the human to learn. Conversely, when the human's initial error is large, accurately estimating how fast they learn becomes important and errors in this estimates may have larger impact on estimating feature value.

%% file: discussion.tex
In this paper, we formally model human decision-making with algorithmic assistance when the human is learning through repeated interactions, and study optimal algorithmic strategies in this context. 
We identify two fundamental tradeoffs in the space of AI-assisted decision making. The first tradeoff, ``informativeness vs. divergence,''  applies to any AI-assisted decision making setting in which the algorithm is designed to choose information for a human's use. 
This tradeoff is amplified in repeated interactions where the human learns from experience. In this case, the algorithm  that determines what the human will learn, faces a second tradeoff between ``fixed vs. growth mindset'' --  optimize with respect to the human's current knowledge or provide them with opportunities for learning, which may take longer but lead to better long-term performance.
We propose a stylized model that unravels how time preferences and the human's ability to learn influence the fixed vs. growth mindset tradeoff. Our results highlight the importance of modeling the human decision-making process and incorporating human learning when designing algorithms to assist decision-makers.

In our modeling, we chose assumptions that are both simple and widely used in the literature, yet sufficient to reveal  fundamental principles in the interaction between algorithmic assistance and human learning. 
For example, we assumed that the variables in linear regression {are drawn from} independent distributions -- a common assumption in learning over time literature (e.g., bandit problems and online learning in general (\cite{slivkins2019introduction})). As future work, it would be interesting to explore correlations between variables or decision instances over time. 
Exploring alternative performance metrics beyond mean squared error and learning models beyond linear regression is also of interest. For example, when performance metrics are discontinuous, such as in binary classification problems, algorithmic decisions may aim to %subtly 
shift the human's prediction to cross a decision threshold, potentially raising ethical considerations regarding manipulation.
An additional direction is to study a human decision maker who recognizes that the algorithm is optimizing and might not provide the features they expect. In this case, the human might act strategically according to their beliefs, leading to a game setting.

We believe that the phenomena we identify extend beyond our specific model, capturing fundamental principles of human-algorithm interactions. These insights lay the groundwork for further research, advancing our understanding of how algorithmic assistance interacts with human learning.

%% file: appendix.tex
\section{On Standardized Features} \label{app:general}

\begin{lemma} \label{thm:standardized-ftrs}
    Let $\{z_i\}_{i=1}^n$ be independent random variables, with known means $\{\mu_i\}_{i=1}^n$ and known finite standard deviations $\{\sigma_i\}_{i=1}^n$, and let $y = a_0 + \sum_{i=1}^n a_i z_i$ be a linear combination of the features that we are willing to represent, for $a_i \in 	\mathbb{R}$ for all $i \in [n]$. Then, w.l.o.g., we can assume $\mu_i = 0$ and $\sigma_i = 1$ for all $i \in [n]$.
\end{lemma}

\begin{proof}
    For all $i$, we use $x_i(z_i)=(z_i - \mu_i) / \sigma_i$. Therefore, the distribution of $x_i$ has zero mean and unit variance. 
    We use coefficients $a'_i$ as follows: $a'_0 = a_0 + \sum_{i=1}^n \mu_i a_i$, and for all $i \in [n]$, $a'_i = \sigma_i a_i$.
    Now, we have: 

    $$y =  a_0 + \sum_{i=1}^n a_i z_i = a_0 + \sum_{i=1}^n \mu_i a_i + \sum_{i=1}^n \sigma_i \cdot a_i (z_i - \mu_i) / \sigma_i = a'_0 + \sum_{i=1}^n a'_i x_i = y'$$
    
    Thus, when the means and standard deviations of the $z_i$'s are known, using the standardized features is simply a change of variables that does not change the problem.
    
\end{proof}

\section{Figure for the Example in Section \ref{sec:example-learning}} \label{app:fig-theory-fit-simulation}

\begin{figure}[ht]
        \centering
\includegraphics[width=0.5\textwidth]{"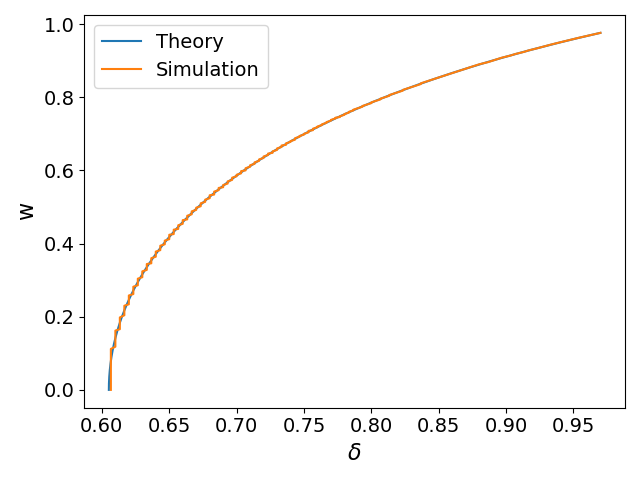"}

    % \hfill
    \caption{The transition curve for the example in Section \ref{sec:example-learning}: For $\delta$ values to the right of this curve, selecting the more informative feature (of the two) is preferred, while for $\delta$ values to the left of this curve, the less divergent feature is preferred. 
    The plot demonstrates that the empirical curve in Figure \ref{fig:w-delta-heatmap} coincides with the theoretical curve derived in \cref{clm:educating-two-features}.
    }
    \label{fig:theory-fit-simulation}
\end{figure}

\FloatBarrier